\documentclass{article} %
\usepackage{iclr2026_conference,times}
\iclrfinalcopy

\usepackage{amsmath,amsfonts,bm}

\def\eqref#1{equation~\ref{#1}}

\def\1{\bm{1}}

\DeclareMathAlphabet{\mathsfit}{\encodingdefault}{\sfdefault}{m}{sl}
\SetMathAlphabet{\mathsfit}{bold}{\encodingdefault}{\sfdefault}{bx}{n}

\usepackage{hyperref}
\usepackage{url}
\usepackage{booktabs}
\usepackage{amsmath, amssymb}

\usepackage{caption}
\usepackage{subcaption}
\usepackage{graphicx}

\usepackage{tikz}
\usetikzlibrary{tikzmark, shapes.geometric, arrows.meta, fit}

\PassOptionsToPackage{table}{xcolor}
\usepackage[table]{xcolor}
\colorlet{lgreen}{green!10}
\colorlet{lblue}{blue!10}
\colorlet{lred}{red!10}

\usepackage{listings}
\definecolor{Numbers}{RGB}{0,137,84}      %
\definecolor{Definitions}{RGB}{0,0,255}   %
\definecolor{Functions}{RGB}{126,93,23}   %
\definecolor{Variables}{RGB}{0,17,134}    %
\definecolor{Comments}{RGB}{0,100,0}      %
\definecolor{Strings}{RGB}{178,0,2}       %
\definecolor{Operators}{RGB}{0,0,0}       %
\definecolor{BaseFunctions}{RGB}{192,0,227} %
\definecolor{Bakgroung}{RGB}{255,255,255} %

\lstdefinelanguage{PythonVSCodeLight}{
    language=Python,
    morekeywords={
        torch, nn, autograd, optim, utils, cuda, device, Tensor,
        tensor, from_numpy, zeros, ones, randn, manual_seed, no_grad, save, load,
        Linear, Conv2d, ReLU, Sigmoid, Tanh, Dropout, Sequential, Module,
        SGD, Adam, lr_scheduler,
        CrossEntropyLoss, MSELoss, BCELoss, NLLLoss,
    },
    keywords=[2]{abs, norm, kthvalue, where},
    keywordstyle=[2]\color{Functions}, %
    sensitive=true
}

\usepackage{tikz}
\NewDocumentCommand{\progressbar}{m O{2} O{red!70} O{0.3}}{%
  \begin{tikzpicture}
    \fill[gray!30] (0,0) rectangle (#2,#4); %
    \fill[#3] (0,0) rectangle ({#1*#2},#4); %
  \end{tikzpicture}%
}

\title{TTQ: Activation-Aware Test-Time Quantization to Accelerate LLM Inference On The Fly}

\author{Toshiaki Koike-Akino, Jing Liu, Ye Wang 
\\
Mitsubishi Electric Research Laboratories (MERL)\\
201 Broadway, Cambridge, MA 02139, USA \\
}

\begin{document}

\maketitle

\begin{abstract}
To tackle the huge computational demand of large foundation models, activation-aware compression techniques without retraining have been introduced. 
However, since these methods highly rely on calibration data, domain shift issues may arise for unseen downstream tasks.
We propose a test-time quantization (TTQ) framework which compresses large models on the fly at inference time to resolve this issue.
With an efficient online calibration, instant activation-aware quantization can adapt every prompt regardless of the downstream tasks, yet achieving inference speedup. 
Several experiments demonstrate that TTQ can improve the quantization performance over state-of-the-art baselines.
\end{abstract}

\section{Introduction}

Large foundation models~\cite{touvron2023llama, achiam2023gpt,liu2023LLaVA} have shown excellent performance across a variety
of  tasks~\citep{wei2022emergent,katz2024gpt, bubeck2023sparks}. 
Nonetheless, these models, with billions of parameters, demand significant computational resources~\citep{schwartz2020green}. 
Towards increasing the accessibility of large language models (LLMs), a number of compression methods~\cite{xu2023survey, zhu2024survey, bai2024beyond} have been introduced: e.g., partial activation~\citep{jiang2024mixtral, lin2024moe}, weight pruning~\citep{frantar2023sparsegpt, sun2023simple, bai2024sparsellm, hassibi1993optimal},
quantization~\citep{frantar2022gptq, lin2024awq, wang2024q}, knowledge distillation~\citep{hsieh2023distilling,  hwang2024pc}, and rank reduction~\citep{yuan2023asvd, liu2024deepseek, hwang2024pc, saxena2024eigen}. 
Related literature is further discussed in Appendix~\ref{sec:related}.

\textit{Test-time scaling}~\citep{chen2024expanding, muennighoff2025s1} is a new paradigm to improve LLM performance by increasing inference computation. 
Instead of increasing the complexity, \textit{test-time compression} is aimed at reducing the total cost of inference on the fly. 
For instance, test-time pruning has been used as a mixture-of-expert (MoE) framework, which dynamically selects important modules depending on each prompt.
Recently, micro-grained MoE~\citep{koike2025mu} exploits a fast activation-aware weight pruning method at test-time to accelerate LLMs.
However, unstructured pruning generally does not improve the hardware efficiency unlike quantization or rank reduction.
We hence propose a new \textbf{test-time quantization (TTQ)} framework, enabled by a fast activation-aware quantization method.
We make the following contributions: (1) Our TTQ accelerates LLMs at inference time, (2) introducing low-complexity activation-aware quantization to compress LLMs on the fly, with negligible overhead, and (3) low-rank decomposition integrated into TTQ, (4) while avoiding domain shift inherent to offline calibration of baseline static quantization. (5) We demonstrate the benefit of TTQ over state-of-the-art methods for several LLM benchmarks.

\section{Test-Time Quantization: TTQ}

\paragraph{Groupwise Quantization}

Round-to-nearest (RTN) is the simplest quantization method to minimize approximation error:
$    \mathcal{L}_0 = \| W - \hat{W} \|^2$,
where $W\in\mathbb{R}^{d'\times d}$ is a weight matrix and $\hat{W}$ is its quantized version.
RTN uses the groupwise quantization-dequantization (QDQ) operation: $\hat{W} = \mathcal{Q}[W] \triangleq \mathcal{G}^-[\mathcal{G}[W]]$. 
The quantization $\mathcal{G}[\cdot]$ and dequantization $\mathcal{G}^-[\cdot]$ are defined as:
\begin{align}
    W_\mathrm{int} &= \mathcal{G}(W) \triangleq
    \mathsf{round}[
    \mathsf{clamp}_q[
    (W - Z) \oslash S
    ]],
    \qquad
    \hat{W} =\mathcal{G}^-(W_\mathrm{int})
    \triangleq 
    W_\mathrm{int} \circ S + Z,
    \label{eq:qdq}
\end{align}
where $S\in\mathbb{R}^{d'\times d}$ and $Z\in\mathbb{R}^{d'\times d}$ are scale and zero-point parameters.
Here, $\oslash$ is element-wise division, $\circ$ is element-wise product, $\mathsf{round}[x]$ gives the closest integer to $x$, and
$\mathsf{clamp}_q[x]=\min(\max(x, 0), 2^q-1)$ is $q$-bit limiting operator.
See Appendix~\ref{sec:rtn} for more details.
Pseudo-code of RTN with a groupsize $g$ is as follows:
\begin{lstlisting}
def rtn(W, q, g): # q: bits, g: groupsize 
    ddash, d = W.shape # W: (d', d)
    W = W.reshape(-1, g) # grouping (d'*d/g, g)
    Wmax, Wmin = W.amax(axis=1), W.amin(axis=1) # (d'*d/g,)
    S, Z = (Wmax - Wmin) / (2**q - 1), Wmin # scale, zero-point
    Wint = ((W - Z[:, None]) / S[:, None]).round().clamp(0, 2**q - 1) 
    What = Wint * S[:, None] + Z[:, None] # dequantization
    return What.reshape(ddash, d) # reshaping back
\end{lstlisting}
The recent NVFP format~\citep{egiazarian2025bridging} uses a microscaling groupsize of $16$ to accelerate GPU computing.
Most literature claims 2 to 4-fold speedup with 4-bit LLMs~\cite{ frantar2025marlin}.

\paragraph{AWQ: Activation-Aware Quantization}

To improve over na\"{i}ve RTN quantization, the activation-aware framework~\citep{lin2023awq, frantar2022gptq, liu2025awp} leverages activation statistics.
Let $X\in\mathbb{R}^{d\times T}$ be an input activation with embedding dimension $d$ and token length $T$.
The aim is to minimize the approximation loss:
\begin{align}
    \mathcal{L} &\triangleq
    \mathbb{E}_{X}
    \big[
    \big\| (W - \hat{W}) X \big\|^2
    \big]
    =
    \mathrm{tr}[ 
    (W-\hat{W})
    \mathbb{E}_X[XX^\top]
    (W-\hat{W})^\top
    ]
    =
    \|(W-\hat{W}) C^{1/2}\|^2,
    \label{eq:loss}
\end{align}
where $C\triangleq\mathbb{E}_X[XX^\top]\in\mathbb{R}^{d\times d}$ is the auto-correlation statistics of input $X$.
Since $C$ is not exactly known at test time, we estimate it with small amount of calibration data, e.g., via shrunk estimator:
$
    C_\lambda = (1-\lambda) 
    {X} {X}^\top + \lambda \eta I
    $,
where $\lambda$ is a shrinkage parameter~\citep{ledoit2004well}, and $\eta=\|{X}\|^2/d$.
GPTQ~\citep{frantar2022gptq} uses the greedy method inspired by optimal brain surgeon~\citep{hassibi1993optimal}.
It requires the Cholesky factorization, whose complexity is at least of cubic order: $\mathcal{O}[d^3 + d d' T]$.
Whereas, AWQ~\citep{lin2023awq} greatly simplifies the problem by approximating  with a diagonal correlation: 
$ C_\lambda \simeq D \triangleq \mathsf{diag}[{X} {X}^\top + \lambda I]^{\alpha} $,  
where $\mathsf{diag}[C]\triangleq C\circ I$ offers a diagonal matrix, and $\alpha$ is an auxiliary parameter.
Note that the above expression gives the $\ell_2$-norm diagonal $D_{i,i} = (\|{X}_{i,:}\|_2^2+\lambda)^{\alpha}$, while the original AWQ uses $\ell_1$-norm $D_{i,i} = (\|{X}_{i,:}\|_1^2 + \lambda)^{\alpha}$.
Given a diagonal correlation matrix, the closed-form solution for (\ref{eq:loss}) is given by the scaled QDQ operation:
$\hat{W} = \mathcal{Q}[W D^{1/2}] D^{-1/2}$.
The pseudo-code of the AWQ concept is given as follows:
\begin{lstlisting}
def awq(X, W, q, g, p, lam, alpha): # q: bits, g: groupsize, p: lp-norm
    D = (X.norm(p=p, axis=1) + lam) ** alpha # X: (d, T); D: (d,)
    What = rtn(W * D[None, :], q, g) # scaled QDQ
    return What * D.reciprocal()[None, :] # scaling back
\end{lstlisting}
Here, we generalized to any $\ell_p$-norm with arbitrary $p$.
See Appendix~\ref{sec:awq} for more details.

\paragraph{TTQ: Test-Time Quantization with Online AWQ}

\begin{figure}[t]
    \centering
    \includegraphics[width=0.9\linewidth]{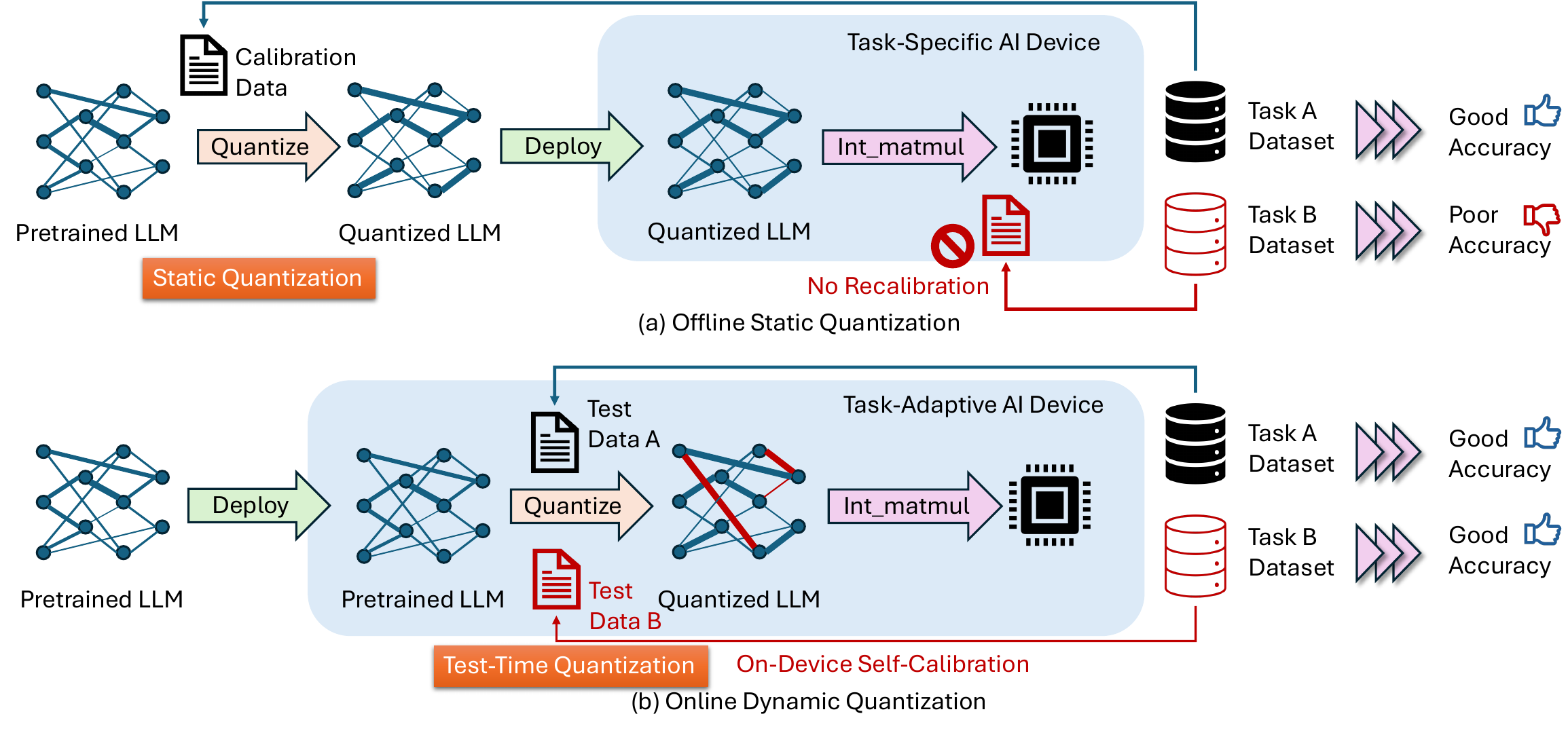}
    \caption{(a) Offline static quantization (e.g., AWQ/GPTQ) requires calibration data, incurs domain shift risk, and cannot be recalibrated after deployment. 
    (b) Our TTQ is online dynamic quantization, with zero offline calibration, and capable of on-device self-calibration at inference time.}
    \label{fig:ttq}
\end{figure}

Activation-aware post-training zero-short quantization like GPTQ/AWQ has several drawbacks:
(1) calibration and extra data are required before deployment; 
(2) a severe domain shift issue may arise at test time when the calibration data were irrelevant; 
and (3) the original full-precision weights are not recoverable for re-calibrating new domain once the quantized model is deployed. 
To solve these issues, we propose a test-time quantization (TTQ) framework shown in Fig.~\ref{fig:ttq}, which employs light-weight online AWQ at inference time.
Specifically, we dynamically calculate the diagonal correlation $D$ for incoming tokens $X$ on the fly to adapt scale $S$ and zero-point $Z$. 
It thus requires \textbf{zero} offline calibration data before deployment. 
Once the weight is quantized, \texttt{int\_matmul} kernels
\footnote{e.g., \texttt{awq\_gemm} CUDA kernel in \url{https://github.com/vllm-project/vllm}} can significantly accelerate the linear projection at GPU devices because of reduced caching overhead.

To keep AWQ fast enough at test time, we keep constant hyperparameters $(\alpha, \lambda, p)$ without exhaustively searching the best values.
Appendix~\ref{sec:param} discusses the hyperparameter selection.
Importantly, the extra computation for online quantization is negligible: specifically, a fraction of overhead complexity over the original un-quantized projection is:
\begin{align}
 \rho &=
    \frac{\mathcal{O}[dT + 3 d' d]}{\mathcal{O}[d'dT]}
    =
    \mathcal{O}\big[ 
    \frac{1}{d'} + \frac{3}{T}
    \big]
    \mathop{\longrightarrow}_{d',T\gg 1}
    0.
\end{align} 
Here, the original projection for $WX$ requires $\mathcal{O}[d'dT]$, the norm calculation for $D_{ii}=(\|X_{i,:}\|^2+\lambda)^\alpha$ requires $\mathcal{O}[dT]$, scaling as $WD^{1/2}$ requires $\mathcal{O}[d'd]$, quantization $\mathcal{G}[\cdot]$ in (\ref{eq:qdq}) needs $\mathcal{O}[d'd]$, and re-scaling back $\mathcal{Q}[WD^{1/2}]D^{-1/2}$ uses another $\mathcal{O}[d'd]$.
Hence, with sufficiently large output dimension $d'\gg 1$ and  token length $T\gg 1$, online AWQ introduces negligible extra-complexity.

\begin{table}[t]
    \centering
    \caption{Calibration length impact at 3-bit quantization with $g=32$ for OPT-350M model. }
    \label{tab:token}    
    \small
    \newcolumntype{g}{>{\columncolor{lgreen}}r}
    \begin{tabular}{l gg rrrrrrr}
    \toprule
    & \multicolumn{2}{c}{\textbf{TTQ}}  & \multicolumn{7}{c}{AWQ (C4 Calib)} \\
    \cmidrule(lr){2-3} \cmidrule(lr){4-10}
    Calib Tokens $T$ & 
    \textbf{0}$_{(r=0)}$  & \textbf{0}$_{(r=16)}$ &
    $2^{11}$ & $2^{12}$ & $2^{13}$ & $2^{14}$ & $2^{15}$ & $2^{16}$ & $2^{17}$     
    \\
    \midrule
    WT2 Perplexity ($\downarrow$) & 
    \underline{24.93} & 
    \textbf{24.17} &
    25.73 &
    25.56 &
    25.62 &
    25.55 &
    25.42 &
    25.07 &
    {25.02} \\
    \bottomrule
    \end{tabular}
\end{table}

\begin{table}[t]
    \centering
    \caption{Groupsize impact on WT2 perplexity at 3-bit quantization  for Qwen3-1.7B model. 
    }
    \label{tab:group}    
    \small
    \setlength{\tabcolsep}{5pt}
    \newcolumntype{g}{>{\columncolor{lgreen}}r}
    \begin{tabular}{l rrrr rrrr}
    \toprule
    Groupsize $g$ & 
    8 & 16  & 32 & 64 & 128 & 256 & 512 & 1024 \\
    \midrule
    RTN &
    33.39 & 66.61 &
    120.89 & 114.97 &
    1450.28 & 15503.10 &
    949478.44 & 57443.80 
    \\
    AWQ (WT2 Calib) & 
    17.64 & \tikzmarknode{s1}{19.00} &
    \tikzmarknode{s2}{20.35} & 20.69 &
    \tikzmarknode{s3}{24.96} & 31.41 &
    36.14 & 68.96 
     \\
    \rowcolor{lgreen}
    \textbf{TTQ} ($r=16$) & 
    \textbf{17.54} & 
    \textbf{17.81} &
    \tikzmarknode{e1}{\textbf{18.29}} &
    \tikzmarknode{e2}{\textbf{19.46}} &
    \textbf{22.33} &
    \tikzmarknode{e3}{\textbf{24.81}}  &
    \textbf{31.43} &
    \textbf{48.70} \\
    \bottomrule
    \end{tabular}
    \begin{tikzpicture}[remember picture, overlay, -{Stealth[scale=1.2]}, red, thick]
        \node[draw, ellipse, fit=(s1), inner sep=0.5pt] (circ) {};
        \draw[->] (circ) -- (e1);
        \node[draw, ellipse, fit=(s2), inner sep=0.5pt] (circ) {};
        \draw[->] (circ) -- (e2);
        \node[draw, ellipse, fit=(s3), inner sep=0.5pt] (circ) {};
        \draw[->] (circ) -- (e3);
    \end{tikzpicture}
\end{table}

\paragraph{TTQ with Low-Rank Decomposition}

Nonetheless, extreme bit quantization often causes a severe degradation, and QLoRA~\citep{dettmers2023qlora} could compensate for quantized LLM performance by adapting low-rank factors.
It motivates us to integrate TTQ with low-rank decomposition. 
We now have low-rank factors with quantized weights:
$    \hat{W} = W_\mathrm{q} + B A$,
where $W_\mathrm{q}\in \mathbb{R}^{d'\times d}$ is quantized residual weights from un-quantized low-rank factors $B\in\mathbb{R}^{d'\times r}$ and $A\in\mathbb{R}^{r\times d}$ with a rank $r$.
When $r\ll \min(d,d')$, it can accelerate the inference as the complexity for low-rank projection is reduced from $\mathcal{O}[d'dT]$ for $WX$ to $\mathcal{O}[r(d'+d)T]$ for $BAX$ --- hence the relative extra-complexity is negligible: $\mathcal{O}[r/d+r/d'] \rightarrow 0$.
A key difference from QLoRA is that TTQ \textbf{dynamically} adapts $W_\mathrm{q}$ depending on $X$, whereas QLoRA uses a \textbf{static} $W_\mathrm{q}$ but adapting only $B$ and $A$.
We may initialize low-rank factors $B$ and $A$ by top-$r$ principal components, and $W_\mathrm{q}$ is obtained on the fly to quantize the residual weights through scaled QDQ: $W_\mathrm{q}=\mathcal{Q}[(W-BA)D^{1/2}]D^{-1/2}$. 
Appendix~\ref{sec:init} discusses some alternative initializations.
Optionally, we may dynamically adapt $B$ and $A$, e.g., through online PCA~\citep{feng2013online}.
Nevertheless, our primary objective is to accelerate LLMs at the inference time through the use of dynamic weight quantization, and thus we focus on static low-rank factors.
Appendix~\ref{sec:runtime} shows up-to \textbf{5-fold} speedup even with an overhead of low-rank projection.

\paragraph{Experiments}

We evaluate the OPT~\citep{zhang2022opt}, Qwen3~\citep{yang2025qwen3}, and Gemma3~\citep{team2025gemma} LLM models on the witkitext-2 (WT2)~\citep{merity2016pointer}, PTB~\citep{marcus1994penn}, and C4~\citep{raffel2020exploring} benchmarks. 
See details of experimental setup in Appendix~\ref{sec:setup}.
We use a groupsize of $g=32$ for all experiments unless specified.
We first show the benefit of our TTQ, which needs \textbf{zero} calibration data, whereas AWQ is sensitive to the size of calibration data, as shown in Table~\ref{tab:token}.
We show the perplexity score for WT2 on OPT-350M, with C4 calibration for AWQ.
TTQ achieves the best over AWQ baselines, where AWQ degrades the performance when fewer calibration tokens are available.
Table~\ref{tab:group} shows the impact of groupsize $g$ for 3-bit quantized Qwen3-1.7B model.
As expected, micro-scaling ($g<32$) achieves good performance, while micro-scaling alone for RTN cannot compete with AWQ/TTQ.
TTQ can tolerate roughly twice larger groupsizes than AWQ, that can halve the required memory for scale $S$ and zero-point $Z$.
Next we show the performance across different models in Table~\ref{tab:perp_short}, where we show the macro average of perplexity over WT2, PTB, and C4 benchmarks.
We observe that few-bit quantization like $q=2$ is more feasible for larger LLMs, and $5$-bit quantization achieves nearly un-quantized performance for most cases.
TTQ achieves the best over all cases, even though AWQ used a long calibration of $T=2^{17}$.
More importantly, AWQ shows fluctuation across different calibration datasets, indicating the potential risk of such \textbf{static} quantization relying heavily on calibration.
We notice that the original un-compressed Gemma3 is significantly poorer for the PTB dataset, and the macro average perplexity can be slightly better with compression occasionally.
See Appendices~\ref{sec:llm}, \ref{sec:vlm} and \ref{sec:vla} for full results in detail including other VLM and VLA benchmarks.

\begin{table*}[t]
\centering
\caption{Perplexity ($\downarrow$) of OPT/Qwen3/Gemma3 models with different quantization methods. 
It shows macro average across WT2/PTB/C4 datasets.
Groupsize is $g=32$ for all cases. Calibration token length is $T=2^{17}$ for AWQ.
\textbf{Bold} and \underline{underline} denote the best and second best, respectively.
Asterisk ``*'' indicates reaching competitive performance to the original un-compressed LLM.
}
\label{tab:perp_short}
\small
\resizebox{0.467\linewidth}{!}{
\begin{tabular}{l rrrr}
  \toprule
  Quantization $q$ & 
  2 bits & 3 bits & 4 bits & 5 bits  
  \\
  \midrule
  \multicolumn{5}{c}{OPT-125M (WT2: 27.7, PTB: 39.0, C4: 26.6, Avg: \textbf{31.1})}
  \\
  \midrule
  RTN & 
  5058.5  & 
  56.3  & 
  33.5  & 
  31.8  
  \\
  AWQ (WT2 Calib) & 
  381.7 &
  37.4 &
  32.3 &
  31.4 
  \\
  AWQ (PTB Calib) &
  375.3 &
  37.3 &
  32.2 &
  31.3 
  \\
  AWQ (C4 Calib) &
  451.7 &
  37.7 &
  32.6 &
  31.3 
  \\
  \rowcolor{lgreen}
  \textbf{TTQ} ($r=0$) &
  \underline{257.4} & 
  \underline{36.6} &  
  \underline{31.9} & 
  \underline{31.2} 
  \\
  \rowcolor{lgreen}
  \textbf{TTQ} ($r=16$) &
  \textbf{141.7} & 
  \textbf{35.8} &  
  \textbf{31.8} &  
  *\textbf{31.1}   
  \\
  \toprule
  \multicolumn{5}{c}{OPT-1.3B (WT2: 14.6, PTB: 20.3, C4: 16.1, Avg: \textbf{17.0})}
  \\
  \midrule
  RTN & 
  11514.4  & 
  27.2  & 
  18.1  & 
  \underline{17.2} 
  \\
  AWQ (WT2 Calib) &
  32.4 &
  \underline{18.0} &
  17.3 &
  *\textbf{17.0} 
  \\
  AWQ (PTB Calib) &
  32.6 &
  18.1 &
  17.3 &
  *\textbf{17.0} 
  \\
  AWQ (C4 Calib) &
  \underline{31.7} &
  \underline{18.0} &
  \underline{17.2} &
  *\textbf{17.0} 
  \\
  \rowcolor{lgreen}
  \textbf{TTQ} ($r=0$) &
  {32.2} & 
  {18.2} & 
  \underline{17.2} & 
  *\textbf{17.0} 
  \\
  \rowcolor{lgreen}
  \textbf{TTQ} ($r=16$) &
  \textbf{27.2} & 
  \textbf{17.9} & 
  \textbf{17.1} & 
  *\textbf{17.0} 
  \\
  \toprule
  \multicolumn{5}{c}{OPT-2.7B (WT2: 12.5, PTB: 18.0, C4: 14.3, Avg: \textbf{14.9})}
  \\
  \midrule
  RTN & 
  6274.5  & 
  36.0  & 
  15.7  & 
  \underline{15.0}  
  \\
  AWQ (WT2 Calib) & 
  23.1  & 
  \underline{15.7}  & 
  \underline{15.1}  & 
  \underline{15.0} 
  \\
  AWQ (PTB Calib) & 
  23.2  & 
  \underline{15.7}  & 
  \textbf{15.0}  & 
  \underline{15.0}   
  \\
  AWQ (C4 Calib) & 
  \underline{22.9}  & 
  \underline{15.7}  & 
  \textbf{15.0}  & 
  \underline{15.0}   
  \\
  \rowcolor{lgreen}
  \textbf{TTQ} ($r=0$) &
  {23.7} & 
  \underline{15.7} &  
  \textbf{15.0} & 
  *\textbf{14.9} 
  \\
  \rowcolor{lgreen}
  \textbf{TTQ} ($r=16$) &
  \textbf{21.2} & 
  \textbf{15.5} &  
  \textbf{15.0} & 
  *\textbf{14.9} 
  \\
  \toprule
  \multicolumn{5}{c}{OPT-6.7B (WT2: 10.9, PTB: 15.8, C4: 12.7, Avg: \textbf{13.1})}
  \\
  \midrule
  RTN & 
  5716.5  & 
  26.2  & 
  13.7  & 
  \underline{13.2}  
  \\
  AWQ (WT2 Calib) &
  17.2 &
  \underline{13.5} &
  \underline{13.2} &
  *\textbf{13.1}
  \\
  AWQ (PTB Calib) &
  17.1 &
  13.6 &
  \underline{13.2} &
  *\textbf{13.1} 
  \\
  AWQ (C4 Calib) &
  \underline{16.9} &
  13.6 &
  \underline{13.2} &
  *\textbf{13.1} 
  \\
  \rowcolor{lgreen}
  \textbf{TTQ} ($r=0$) &
  {17.2} & 
  {13.6} &  
  \underline{13.2} &
  *\textbf{13.1} 
  \\
  \rowcolor{lgreen}
  \textbf{TTQ} ($r=16)$&
  \textbf{16.3} & 
  \textbf{13.4} &  
  *\textbf{13.1} &
  *\textbf{13.1} 
  \\
  \bottomrule
\end{tabular}
}
\hfill
\resizebox{0.52\linewidth}{!}{
\begin{tabular}{l rrrr}
  \toprule
  Quantization $q$ & 
  2 bits & 3 bits & 4 bits & 5 bits  
  \\
  \midrule
  \multicolumn{5}{c}{Qwen3-0.6B (WT2: 21.0, PTB: 43.8, C4: 30.3, Avg:  \textbf{31.7})}
  \\
  \midrule
  RTN & 
  8.2e6  & 
  127.3  & 
  38.2  & 
  33.5  
  \\
  AWQ (WT2 Calib) & 
  9739.1 &
  49.4 &
  33.5 &
  32.4 
  \\
  AWQ (PTB Calib) &
  17344.3 &
  47.9 &
  34.1 &
  \underline{32.0} 
  \\
  AWQ (C4 Calib) &
  5388.1 &
  48.3 &
  \underline{33.0} &
  32.1 
  \\
  \rowcolor{lgreen}
  \textbf{TTQ} ($r=0$) &
  \underline{2827.8} & 
  \underline{44.7} &  
  \underline{33.0} & 
  \textbf{31.9} 
  \\
  \rowcolor{lgreen}
  \textbf{TTQ} ($r=16$) &
  \textbf{1552.6} & 
  \textbf{42.0} &  
  \textbf{32.9} &  
  \textbf{31.9}   
  \\
  \toprule
  \multicolumn{5}{c}{Qwen3-1.7B (WT2: 16.7, PTB: 33.8, C4: 16.1, Avg: \textbf{24.2})}
  \\
  \midrule
  RTN & 
  1.4e6  & 
  162.8  & 
  30.6  & 
  26.1  
  \\
  AWQ (WT2 Calib) &
  {1864.7} &
  30.0 &
  {24.5} &
  {24.4} 
  \\
  AWQ (PTB Calib) &
  2309.6 &
  29.8 &
  {24.7} &
  {24.5} 
  \\
  AWQ (C4 Calib) &
  2364.7 &
  28.2 &
  24.9 &
  {24.4} 
  \\
  \rowcolor{lgreen}
  \textbf{TTQ} ($r=0$) &
  \underline{522.6} & 
  \underline{27.3}  &
  \underline{24.4}    
  &
  \underline{24.3}    
  \\
  \rowcolor{lgreen}
  \textbf{TTQ} ($r=16$) &
  \textbf{264.6} & 
  \textbf{26.4} & 
  \textbf{24.3} &   
  *\textbf{24.1}
  \\
  \toprule
  \multicolumn{5}{c}{Gemma3-270M (WT2: 70.1, PTB: 698.7, C4: 68.1, Avg: \textbf{279.0})}
  \\
  \midrule
  RTN & 
  2.6e11  & 
  1795.0  & 
  391.9  & 
  315.0 
  \\
  AWQ (WT2 Calib) & 
  89188.4 &
  517.4 &
  279.6 &
  306.0
  \\
  AWQ (PTB Calib) &
  1.9e5 &
  537.7 &
  310.0 &
  293.4
  \\
  AWQ (C4 Calib) &
  1.3e5 &
  598.1 &
  326.4 &
  *276.8
  \\
  \rowcolor{lgreen}
  \textbf{TTQ} ($r=0$) &
  \underline{30199.0} & 
  \underline{387.7} &  
  *\underline{277.0} & 
  *\underline{262.5} 
  \\
  \rowcolor{lgreen}
  \textbf{TTQ} ($r=16$) &
  \textbf{19657.0} &
  \textbf{382.8} &
  *\textbf{275.7} &
  *\textbf{261.5}
  \\
  \toprule
  \multicolumn{5}{c}{Gemma3-1B (WT2: 27.7, PTB: 212.4, C4: 33.2, Avg: \textbf{91.1})}
  \\
  \midrule
  RTN & 
  8.6e5  & 
  209.4  & 
  111.1  & 
  96.6  
  \\
  AWQ (WT2 Calib) &
  {4734.7} &
  134.7 &
  {91.9} &
  {95.9} 
  \\
  AWQ (PTB Calib) &
  9326.6 &
  138.9 &
  {99.2} &
  {95.5}
  \\
  AWQ (C4 Calib) &
  5486.9 &
  150.8 &
  93.5 &
  {93.6} 
  \\
  \rowcolor{lgreen}
  \textbf{TTQ} ($r=0$) &
  \underline{1928.5} &
  \underline{127.3} &
  *\textbf{89.9} &
  *\textbf{90.2} 
  \\
  \rowcolor{lgreen}
  \textbf{TTQ} ($r=16$) &
  \textbf{1804.9} &
  \textbf{114.5} &
  \underline{91.7} &
  *\underline{90.3}
  \\
  \bottomrule
\end{tabular}
}
\end{table*}

\section{Conclusion}

We proposed a new TTQ framework, which employs a light-weight activation-aware quantization at inference time to accelerate LLMs, without requiring offline calibration or fine-tuning.
We demonstrated a consistent advantage over state-of-the-art baselines on several LLM models.
Further benchmark evaluations will be reported in the future, and we plan to integrate test-time pruning and decomposition into TTQ. 
How to dynamically adjust hyperparameters is interesting to explore.

\bibliography{ref}
\bibliographystyle{iclr2026_conference}

\appendix
\section{Related Work}
\label{sec:related}

\subsection{Model Compression}
The field of model compression for LLMs has aimed at mitigating the substantial computation and memory requirements~\cite{zhu2024survey, yuan2024llm}. 
Such methods primarily fall into four categories: weight quantization~\cite{lin2024awq, frantar2022gptq, wang2024q}, network pruning~\cite{lecun1989optimal, hassibi1993optimal, frantar2023sparsegpt, bai2024sparsellm},
knowledge distillation~\cite{hsieh2023distilling, hwang2024pc}, and rank reduction~\cite{yuan2023asvd, liu2024deepseek, hwang2024pc, saxena2024eigen, saha2024compressing}. 

\subsection{Dynamic Quantization}
Dynamic quantization is a strategy for adapting precision at runtime rather than using fixed bit-widths across all layers and inputs. 
DNQ~\cite{xu2018dnq} uses a controller that learns per-layer bit-width policies to balance accuracy and compression.
Recent research~\cite{santini2025probabilistic} uses probabilistic frameworks for input-adaptive quantization, where quantization parameters are conditioned on each input, demonstrating negligible performance loss with adaptive schemes compared to static counterparts.

\subsection{Few-Bit Quantization}

Extreme 1-bit quantization~\cite{qin2023bibench} has shown relatively good performance even with 1-bit quantization, e.g., BNN~\cite{courbariaux2016binarized, kim2016bitwise}, XNOR-Net~\cite{rastegari2016xnor}, DoReFa-Net~\cite{zhou2016dorefa}, Bi-Real~\cite{liu2018bi}, IR-Net~\cite{qin2020forward}, RBNN~\cite{lin2020rotated}, and ABC-Net~\cite{lin2017towards}.
They proposed a variety of gradient approximation methods for non-differentiable binarization operation.
For example, BNN uses STE trick for non-differentiable sign function.

Binarized networks have many papers on hardware implementation, mostly on generic field-programmable gate-array (FPGA) platform, e.g., \cite{zhao2017accelerating, zhang2021fracbnn}.
Nevertheless, binarization often causes a substantial loss, and most quantization papers~\cite{frantar2022gptq, lin2023awq} for foundation models consider at least 3 bits to achieve an acceptable performance.
Whereas, 2-bit quantization was shown feasible for LLM compression with vector quantization~\cite{chee2023quip, egiazarian2024extreme}.

\subsection{Activation/Hardware-Aware Quantization}

DeepShift~\cite{elhoushi2021deepshift} uses power-of-two weights to eliminate multiplication operations.
HAWQ~\cite{dong2019hawq} uses layer-wise quantization based on optimal brain pruning~\cite{lecun1989optimal}.
HAWQv2~\cite{dong2020hawq} considers mixed-precision weight and activation.
HAWQv3~\cite{yao2021hawq} uses integer weight in dyadic format.
HEQ~\cite{koike2024hardware} jointly minimizes quantization loss and computing energy on custom hardware.
Then, GPTQ~\cite{frantar2022gptq} extends HAWQ using zero-shot calibration, while activation-aware quantization (AWQ)~\cite{lin2023awq} uses activation-dependent scaling.
GPFQ~\cite{zhang2023post} uses a greedy path-following mechanism to quantize weights with theoretical guarantees.
AWP~\cite{liu2025awp} uses projected gradient descent for activation aware quantization and pruning based on compressed sensing framework.
QEP~\cite{arai2025quantization} mitigates error propagation across layers.
QLoRA~\cite{dettmers2023qlora} uses low-rank adapter to compensate for the loss of quantized LLMs.

\subsection{Test-Time Computing}

Prior work has explored improving performance at inference time by increasing computation, adaptation, or interaction beyond a single forward pass. 
Test-time scaling~\cite{chen2024expanding, muennighoff2025s1, snell2024scaling} allocate additional inference-time compute via sampling, search, and reasoning while keeping model parameters fixed, whereas test-time computing dynamically structures inference through adaptive computation, tool use, and external memory. 
In contrast, test-time update/adaptation~\cite{liang2025comprehensive, chen2022contrastive} performs unsupervised or self-supervised parameter updates during deployment, often under distribution shift, with meta-learning explicitly training models for rapid test-time adaptation. 
Test-time reinforcement learning~\cite{zuo2025ttrl, odena2017changing} further frames inference as an interactive process with online policy improvement driven by reward signals.
$\mu$-MoE~\cite{koike2025mu} is a new test-time pruning framework, using activation-aware pruning on the fly at the inference time.

\section{RTN: Groupwise Round-to-Nearest Quantization}
\label{sec:rtn}

Round-to-nearest (RTN) is the simplest quantization method to minimize approximation error:
\begin{align}
    \mathcal{L}_0 &= \| W - \hat{W} \|^2,
\end{align}
where $W\in\mathbb{R}^{d'\times d}$ is a weight matrix and $\hat{W}$ is its quantized version.
RTN uses the groupwise quantization-dequantization (QDQ) operation: 
\begin{align}
 \hat{W} &= \mathcal{Q}[W] \triangleq \mathcal{G}^-[\mathcal{G}[W]],   
\end{align}
Here, the quantization $\mathcal{G}[\cdot]$ and dequantization $\mathcal{G}^-[\cdot]$ are defined as:
\begin{align}
    W_\mathrm{int} &= \mathcal{G}(W) \triangleq
    \mathsf{round}[
    \mathsf{clamp}_q[
    (W - Z) \oslash S
    ]],
    \\
    \hat{W} &=\mathcal{G}^-(W_\mathrm{int})
    \triangleq 
    W_\mathrm{int} \circ S + Z,
    \label{eq:qdq-app}
\end{align}
where $S\in\mathbb{R}^{d'\times d}$ and $Z\in\mathbb{R}^{d'\times d}$ are scale and zero-point parameters.
Here, $\oslash$ is element-wise division, $\circ$ is element-wise product, $\mathsf{round}[x]$ gives the closest integer to $x$, and
$\mathsf{clamp}_q[x]=\min(\max(x, 0), 2^q-1)$ is $q$-bit limiting operator.
The straightforward choice of scale and zero-point is 
\begin{align}
 S &= (W_{\max} -W_{\min})/(2^q-1)   ,
 \\
 Z&=W_{\min},
\end{align}
while they can be adjusted to improve accuracy slightly.
The QDQ has several variants and non-uniform formats, such as NF4~\citep{dettmers2023qlora}, see Appendix~\ref{sec:qdq}. 
Pseudo-code of RTN with a groupsize $g$ is as follows:
\begin{lstlisting}
def rtn(W, q, g): # q: bits, g: groupsize 
    ddash, d = W.shape # W: (d', d)
    W = W.reshape(-1, g) # grouping (d'*d/g, g)
    Wmax = W.amax(axis=1, keepdim=True) # (d'*d/g, 1)
    Wmin = W.amin(axis=1, keepdim=True) # (d'*d/g, 1)
    S, Z = (Wmax - Wmin) / (2**q - 1), Wmin # scale, zero-point
    Wint = ((W - Z) / S).round().clamp(0, 2**q - 1) 
    What = Wint * S + Z # dequantization
    return What.reshape(ddash, d) # reshaping back
\end{lstlisting}
The total memory will be $qd'd$ bits for $W_\mathrm{int}$ and $d'd/g$ parameters for $S$ and $Z$.
Larger groupsize reduces the total required memory to represent $\hat{W}$, while smaller groupsize can reduce the quantization error.
The recent NVFP format~\citep{egiazarian2025bridging} uses a microscaling groupsize of $16$ to accelerate GPU computing.
Because most GPUs are not fully compatible to arbitrary bitwidth integer operations, the practical advantage for the quantization comes with the reduction of required memory, which also leads to GPU acceleration due to the reduction of caching bottleneck.
Most literature~\cite{lin2023awq, frantar2022gptq, frantar2025marlin} claims 2 to 4-fold speedup with quantized LLMs.

\section{AWQ: Activation-Aware Quantization}
\label{sec:awq}

To improve over na\"{i}ve RTN quantization, the activation-aware framework~\citep{lin2023awq, frantar2022gptq, liu2025awp} leverages activation statistics.
Let $X\in\mathbb{R}^{d\times T}$ be an input activation with embedding dimension $d$ and token length $T$.
The aim is to minimize the approximation loss:
\begin{align}
    \mathcal{L} &\triangleq
    \mathbb{E}_{X}
    \big[
    \big\| (W - \hat{W}) X \big\|^2
    \big]
    \\
    &=
    \mathrm{tr}[ 
    (W-\hat{W})
    \underbrace{
    \mathbb{E}_X[XX^\top]
    }_{C\in\mathbb{R}^{d\times d}}
    (W-\hat{W})^\top
    ]
    \\
    &=
    \|(W-\hat{W}) C^{1/2}\|^2,
    \label{eq:loss-app}
\end{align}
where $C\triangleq\mathbb{E}_X[XX^\top]\in\mathbb{R}^{d\times d}$ is the auto-correlation statistics of input $X$.
Since $C$ is not exactly known at test time, we estimate it with small amount of calibration data, e.g., via shrunk estimator:
\begin{align}
    C_\lambda &= (1-\lambda) 
    {X} {X}^\top + \lambda \eta I,
\end{align}
where $\lambda$ is a shrinkage parameter, and $\eta=\|{X}\|^2/d$.
For large dimension $d$, the Ledoit--Wolf shrinkage~\citep{ledoit2004well} is reliable.
Note that this is equivalent to minimize the weighted errors of activation-aware and weight-only quantization objectives:
\begin{align}
    \mathcal{L}' &\triangleq
    (1-\lambda) \|(W-\hat{W}) {X} \|^2
    +
    \lambda \eta \| W-\hat{W} \|^2
    \\
    &=
    \|(W-\hat{W})C_\lambda^{1/2}\|^2.
    \label{eq:lprime-app}
\end{align}
Also note that the solution is invariant to any scaling factor to the correlation, e.g., many literature use 
\begin{align}
C'_{\lambda'} &\triangleq 
C_\lambda / (1-\lambda) \\
&= {X}{X}^\top + \lambda' I,    
\end{align}
where $\lambda'=\lambda\eta/(1-\lambda)$ is a so-called damping factor.
We treat any scaled correlation matrix estimate as $C_\lambda$.

When there is no structure in correlation matrix $C_\lambda$, it has no closed-form solution to minimize $\mathcal{L}'$, whereas GPTQ~\citep{frantar2022gptq} uses the greedy method based on the inverse Hessian inspired by optimal brain surgeon~\citep{hassibi1993optimal}.
GPTQ requires the Cholesky factorization, whose complexity is at least of cubic order: $\mathcal{O}[d^3 + d d' T]$.
In addition, it needs extra computations for off-diagonal error cancellation with the Gaussian elimination.
Alternatively, AWP~\citep{liu2025awp}  uses projected gradient descent to solve it iteratively, i.e., at the $k$th step:
\begin{align}
    \hat{W}^{(k+1)} &=
    \mathcal{Q}[
    \hat{W}^{(k)}
    +
    \mu 
    (W - \hat{W}^{(k)}) C_\lambda
    ]
    ,
\end{align}
where $\mu$ is the stepsize. 
It requires a cubic complexity of $\mathcal{O}[d'd^2K]$ for $K$ total iterations.

AWQ~\citep{lin2023awq} greatly simplifies the problem by approximating  with a diagonal correlation: 
\begin{align}
 C_\lambda \simeq D \triangleq \mathsf{diag}[{X} {X}^\top + \lambda I]^{\alpha} ,  
 \label{eq:lp2-app}
\end{align}
where $\mathsf{diag}[C]\triangleq C \circ I$ offers a diagonal matrix zeroing out off-diagonal elements of an argument matrix $C$, and $\alpha$ is an auxiliary parameter to compensate for the diagonal approximation loss.
Note that the above expression gives the $\ell_2$-norm diagonal $D_{i,i} = (\|{X}_{i,:}\|_2^2+\lambda)^{\alpha}$, while the original AWQ uses $\ell_1$-norm $D_{i,i} = (\|{X}_{i,:}\|_1^2 + \lambda)^{\alpha}$.
Given a diagonal correlation matrix, the closed-form solution for (\ref{eq:lprime-app}) is given by the scaled QDQ operation:
\begin{align}
    \hat{W} &= \mathcal{Q}[W D^{1/2}] D^{-1/2}.
\end{align}
Importantly, the diagonal correlation also reduces the computation for $D$ from $\mathcal{O}[d^2T']$ to $\mathcal{O}[dT']$.
AWQ then requires only quadratic extra-complexity of $\mathcal{O}[2dd' + dT']$, yet achieves performance competitive with GPTQ.
The pseudo-code of the AWQ concept is given as follows:
\begin{lstlisting}
def awq(X, W, q, g, p, lam, alpha): # q: bits, g: groupsize, p: lp-norm
    D = X.norm(p=p, axis=1) # X: (d, T)
    D = (D + lam) ** alpha  # D: (d,)
    What = rtn(W * D[None, :], q, g) # scaled QDQ
    return What * D.reciprocal()[None, :] # scaling back
\end{lstlisting}
Here, we generalized to any $\ell_p$-norm with arbitrary $p$.
The hyper-parameters such as $\alpha$ and $p$ can be adjusted by searching for the best values to minimize (\ref{eq:lprime-app}).

\section{Quantization-Dequantization (QDQ) Formats}
\label{sec:qdq}

There are many variants for QDQ operations, where we consider one example:
\begin{align}
    \mathcal{G}(W) &\triangleq
    \mathsf{round}[\mathsf{clamp}_q[
    (W - Z) \oslash S
    ]],
    \\
    \mathcal{G}^-(W_\mathrm{int})
    &\triangleq 
    W_\mathrm{int} \circ S + Z,
\end{align}
where $\mathcal{G}[\cdot]$ and $\mathcal{G}^-[\cdot]$ denote quantization and dequantization operations, respectively.
Instead, many literature use an alternative representation:
\begin{align}
    \mathcal{G}'(W) &\triangleq
    \mathsf{clamp}_q[
    \mathsf{round}[
    W \oslash S'
    ]
    + Z'
    ],
    \\
    \mathcal{G}'^-(W_\mathrm{int})
    &\triangleq 
    (W_\mathrm{int} - Z') \circ S'.
\end{align}
Both representations have no significant difference, and thus we focus on the first one.
Yet another format includes nonuniform scaling like the NF4 format, which is based on the normal distribution.

We consider the asymmetric format for the scale and zero-point:
\begin{align}
    S &= (W_{\max} - W_{\min}) / (2^q - 1),
    \\
    Z &= W_{\min}.
\end{align}

Some literature further adjust the scale and zero-point factors to minimize the quantization error. 
For example, we may use expanded $W'_{\max}$ and $W'_{\min}$ instead of $W_{\max}$ and $W_{\min}$:
\begin{align}
    W'_{\max} & \triangleq 
    \frac{1+\nu}{2} W_{\max} +
    \frac{1-\nu}{2} W_{\min}
    \\
    W'_{\min} & \triangleq 
    \frac{1-\nu}{2} W_{\max} +
    \frac{1+\nu}{2} W_{\min}
\end{align}
where $\nu\in\mathbb{R}$ is an expansion factor.
Note that $\nu=1$ reduces to the standard scaling. 
The best expansion factor is often around $\nu\simeq 0.95$.

Alternatively, the symmetric format is often used:
\begin{align}
    S &= 2 |W|_{\max} / (2^q - 1),
    \\
    Z &= - |W|_{\max}.
\end{align}
As it has fewer degrees of freedom (i.e., $Z$ or $S$ is redundant as $S=-Z/(2^q-1)$), it offers a degraded accuracy in general, while it gains fewer memory requirement.

Yet another option is to extend test-time activation-aware quantization towards vector quantization~\cite{egiazarian2024extreme} and compression~\cite{frantar2023qmoe}.

\section{Quantization-Aware Low-Rank Factorization}
\label{sec:init}

Initialization of low-rank factors may impact the quantization performance.
What is the best low-rank factors $B$ and $A$, to reduce the quantization error of the residual?
The most na\"{i}ve way is to use the top-$r$ principal components of $W$:
\begin{align}
    U_r \Lambda_r V_r & = \mathsf{svd}_r[ W ],
    \\
    B &= U_r \Lambda_r^{1/2}, 
    \\
    A &= \Lambda_r^{-1/2} V_r,
\end{align}
where $\mathsf{svd}_r[\cdot]$ is rank-$r$ truncated singular-value decomposition (SVD), $U_r\in\mathbb{R}^{d'\times r}$, $\Lambda_r\in\mathbb{R}^{r\times r}$, $V_r\in\mathbb{R}^{r\times d}$ are left singular vectors, diagonal singular-values matrix, and right singular vectors, respectively.
If we have a calibration data, we can use ASVD~\cite{yuan2023asvd, koike2025latentllm} instead: $\mathsf{svd}_r[W C^{1/2}_\lambda] C^{-1/2}_\lambda$.

However, it does not guarantee that the residual $W-BA$ can reduce the quantization error in the end.
To consider quantization-aware low-rank factorization, we may use alternating method, e.g., at the $k$th step:
\begin{align}
    B^{(k)} A^{(k)} &= \mathsf{svd}_r[W - W_\mathrm{q}^{(k)}], \\
    W_\mathrm{q}^{(k+1)} &= \mathcal{Q}[W-B^{(k)} A^{(k)}],
\end{align}
with a proper initialization.
This quantization-aware low-rank factorization may reduce the quantization error, while it still has no guarantee that the loss is small at inference time.
We found that the alternating solution had almost no gain, and we focus on the straightforward principal components for static low-rank factor initialization.

\paragraph{Low-Rank Factor Quantization}
Alternatively, the low-rank factors $A$ and/or $B$ can be also quantized online or offline, to further accelerate inference.
For example, we may consider: (1) $A$ is quantized while $B$ is un-quantized; (2)
$B$ is quantized while $A$ is un-quantized; and (3) both $A$ and $B$ are quantized.
The cases (1) and (2) may be more advantageous because the product of $BA$ is no longer quantized.

\paragraph{Low-Rank Factor Pruning}
In addition, we can integrate with pruning with quantization and low-rank factorizations. 
For example, we can use test-time pruning used in $\mu$-MoE~\cite{koike2025mu} in conjunction with TTQ.
Because both uses similar diagonal correlation matrix, we do not need extra computation for $D$. 
Similarly, we can consider pruning for low-rank factors as well: (1) $A$ is pruned while $B$ is un-pruned; (2)
$B$ is pruned while $A$ is un-pruned; and (3) both $A$ and $B$ are pruned.
Because the product of $BA$ is no longer sparse, the cases (1) and (2) may be more advantageous.

\paragraph{Test-Time Decomposition}
Another option is to use test-time low-rank decomposition for adaptive $A$ and $B$ at inference time.
We may consider using efficient online PCA algorithms:
\begin{itemize}
    \item \textbf{Stochastic gradient methods.} Oja’s rule \citep{oja1982simplified} performs stochastic gradient ascent on the Rayleigh quotient to estimate the top principal components from streaming data. 
    
    \item \textbf{Incremental SVD methods.} Deterministic approaches to update low-rank factorizations include incremental eigen-analysis \citep{hall1998incremental} and incremental SVD techniques \citep{brand2002incremental}.
    
    \item \textbf{Subspace tracking algorithms.} Recursive least-squares formulations such as PAST \citep{yang1995projection} provide fast convergence and strong tracking capability under non-stationary data.
    
    \item \textbf{Online optimization perspectives.} Analyzing online PCA via stochastic optimization frameworks establishes finite-sample convergence and performance guarantees \citep{allen2017first}.
\end{itemize}

\section{Hyperparameter Selection for TTQ}
\label{sec:param}

In static/original AWQ, quantization hyperparameters like $\alpha$ can be adjusted at the offline calibration phase.
However, TTQ needs to re-compute scale and zero-points $S$ and $Z$ based on diagonal correlation $C_\mathrm{\lambda}$, dynamically given every incoming tokens $X$.
Hence, searching those hyperparameters on the fly should be restricted.

Nevertheless, if we still can use post-training calibration data, then we can find the best possible hyperparameters and prior correlation matrix in more exhaustive way before test time.
And, online update of correlation matrix is carried out at inference time to improve the correlation estimation accuracy.
In our work, we use constant hyperparameters manually selected.
How to choose the hyperparameters?

\begin{figure}[t]
    \centering
    \begin{subfigure}[b]{0.32\textwidth}
         \centering
         \includegraphics[width=\textwidth]{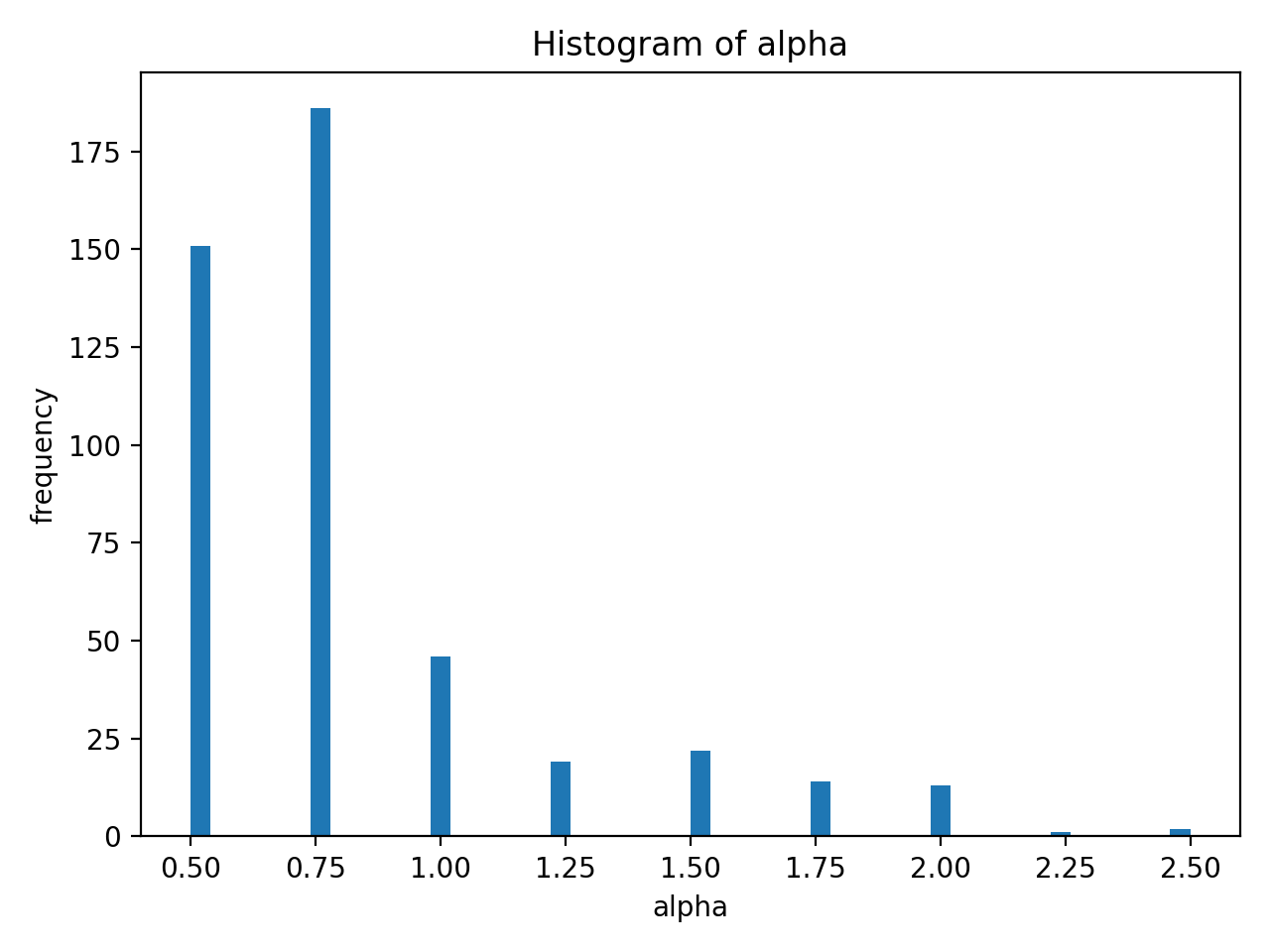}
         \caption{Top-5 $\alpha$}
         \label{fig:best_alpha}
    \end{subfigure}
    \hfill
    \begin{subfigure}[b]{0.32\textwidth}
         \centering
         \includegraphics[width=\textwidth]{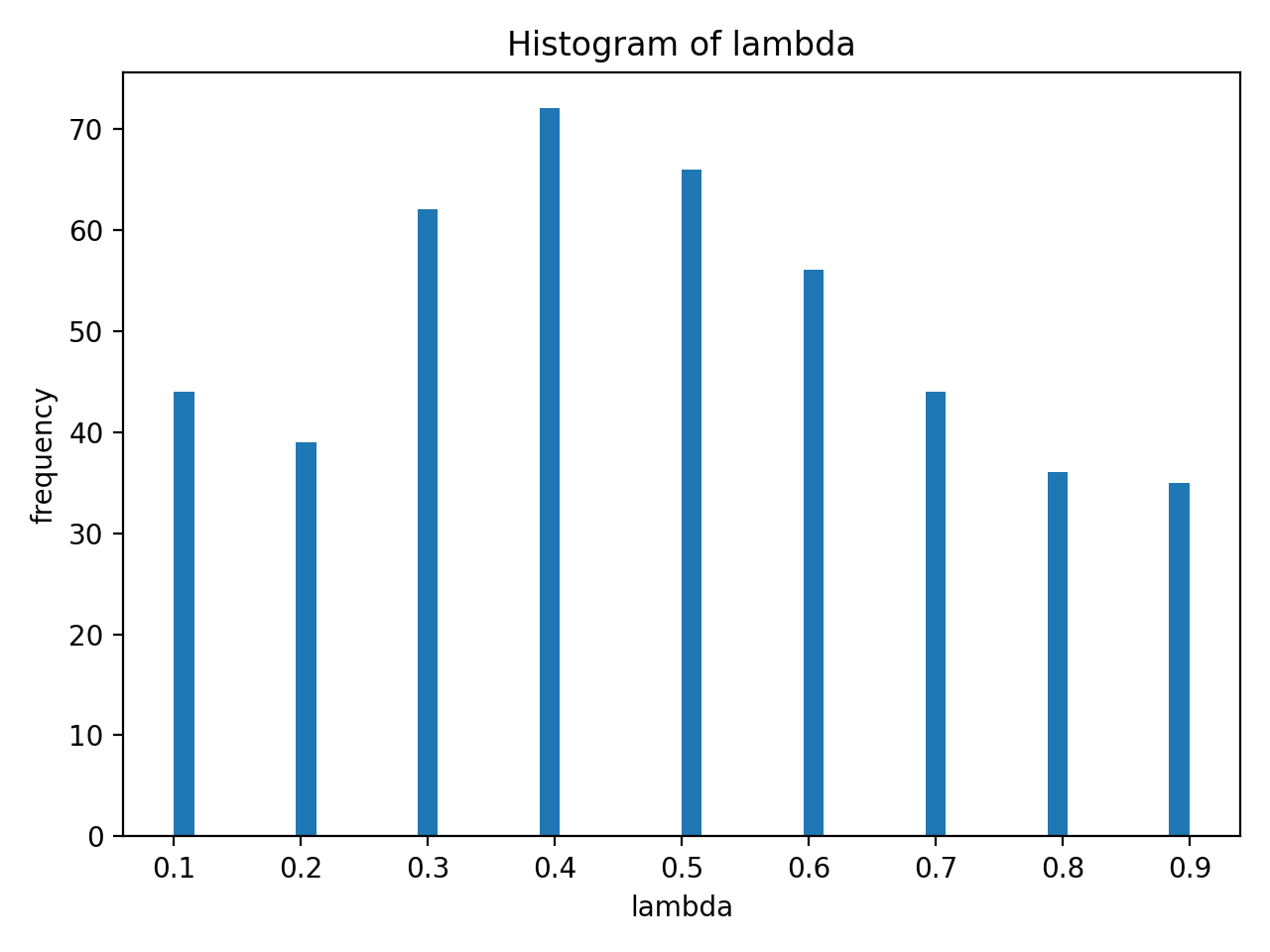}
         \caption{Top-5 $\lambda$}
         \label{fig:best_lambda}
    \end{subfigure}
    \hfill
    \begin{subfigure}[b]{0.32\textwidth}
         \centering
         \includegraphics[width=\textwidth]{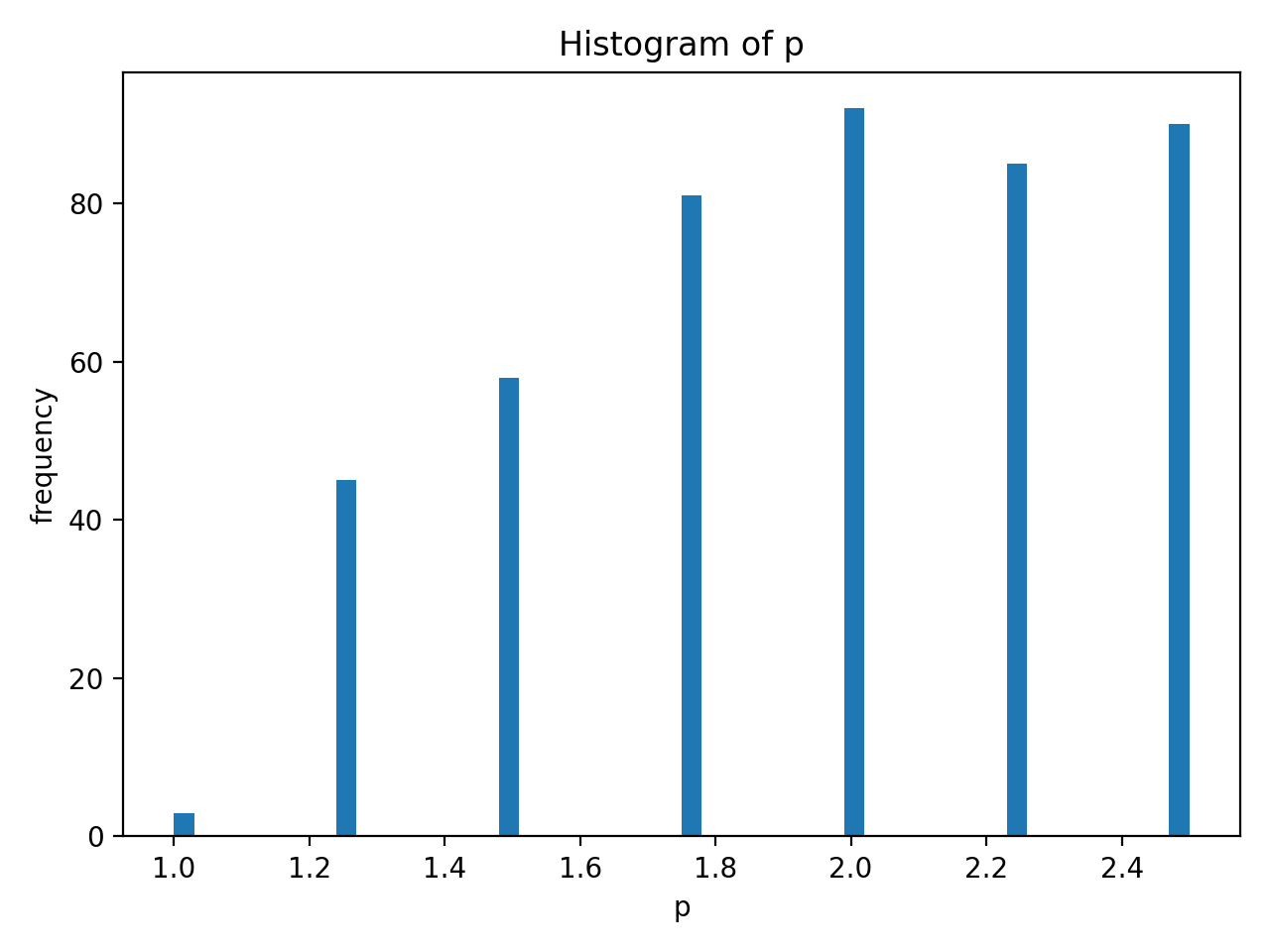}
         \caption{Top-5 $p$}
         \label{fig:best_p}
    \end{subfigure}
    \\
    \begin{subfigure}[b]{0.32\textwidth}
         \centering
         \includegraphics[width=\textwidth]{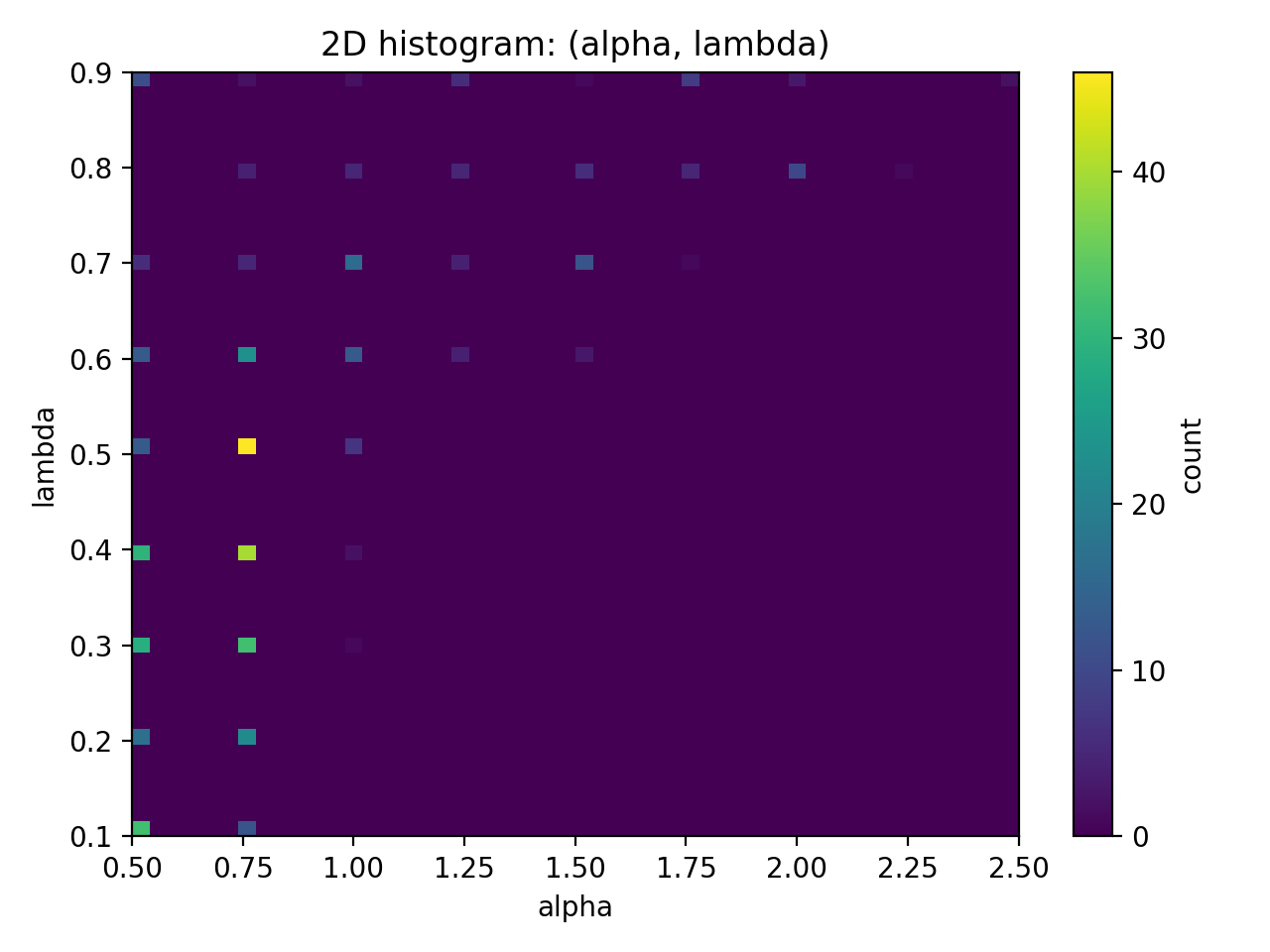}
         \caption{Top-5 $(\alpha, \lambda)$}
         \label{fig:best_alpha_lambda}
    \end{subfigure}
    \hfill
    \begin{subfigure}[b]{0.32\textwidth}
         \centering
         \includegraphics[width=\textwidth]{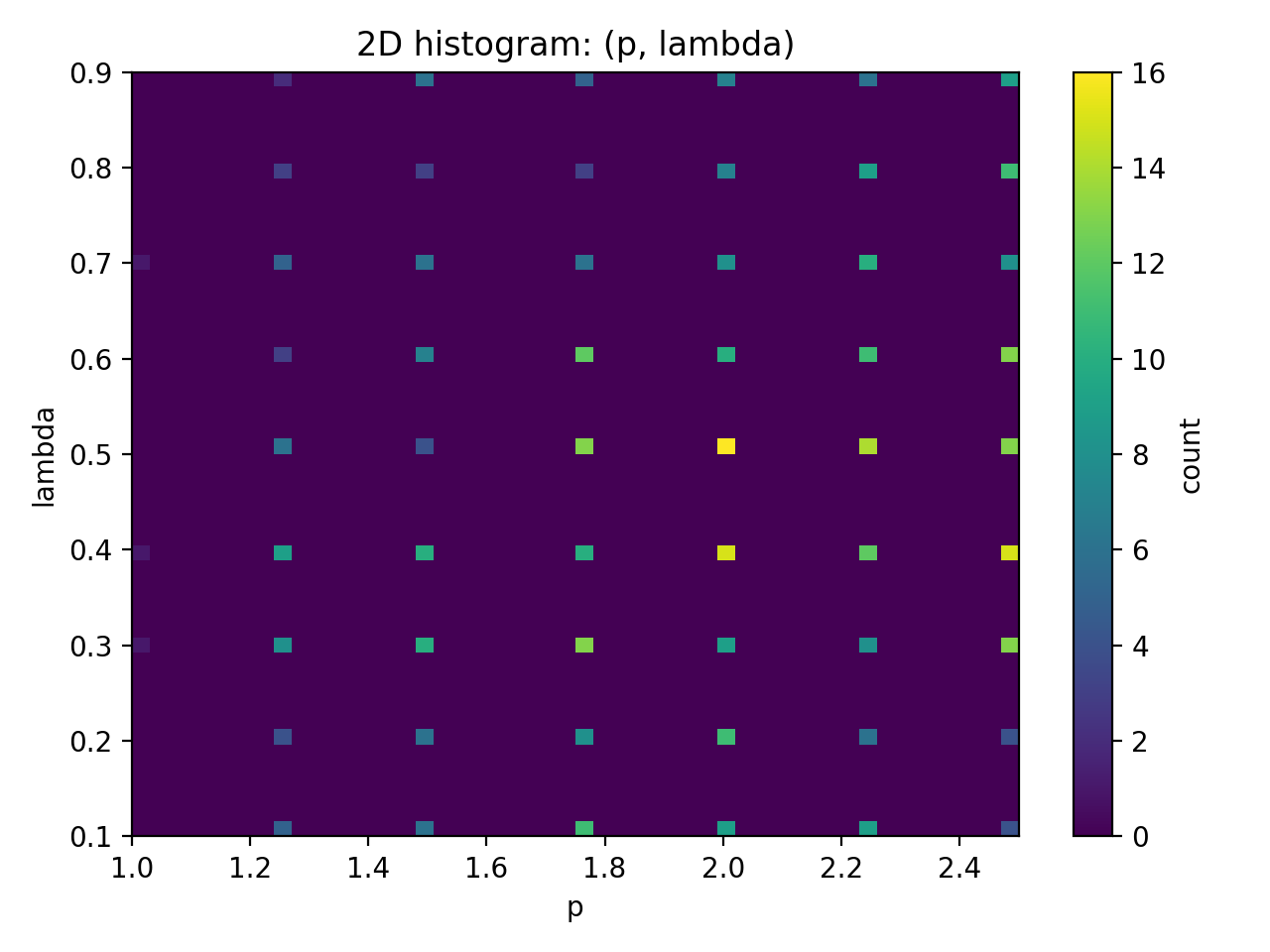}
         \caption{Top-5 $(\lambda, p)$}
         \label{fig:best_p_lambda}
    \end{subfigure}
    \hfill
    \begin{subfigure}[b]{0.32\textwidth}
         \centering
         \includegraphics[width=\textwidth]{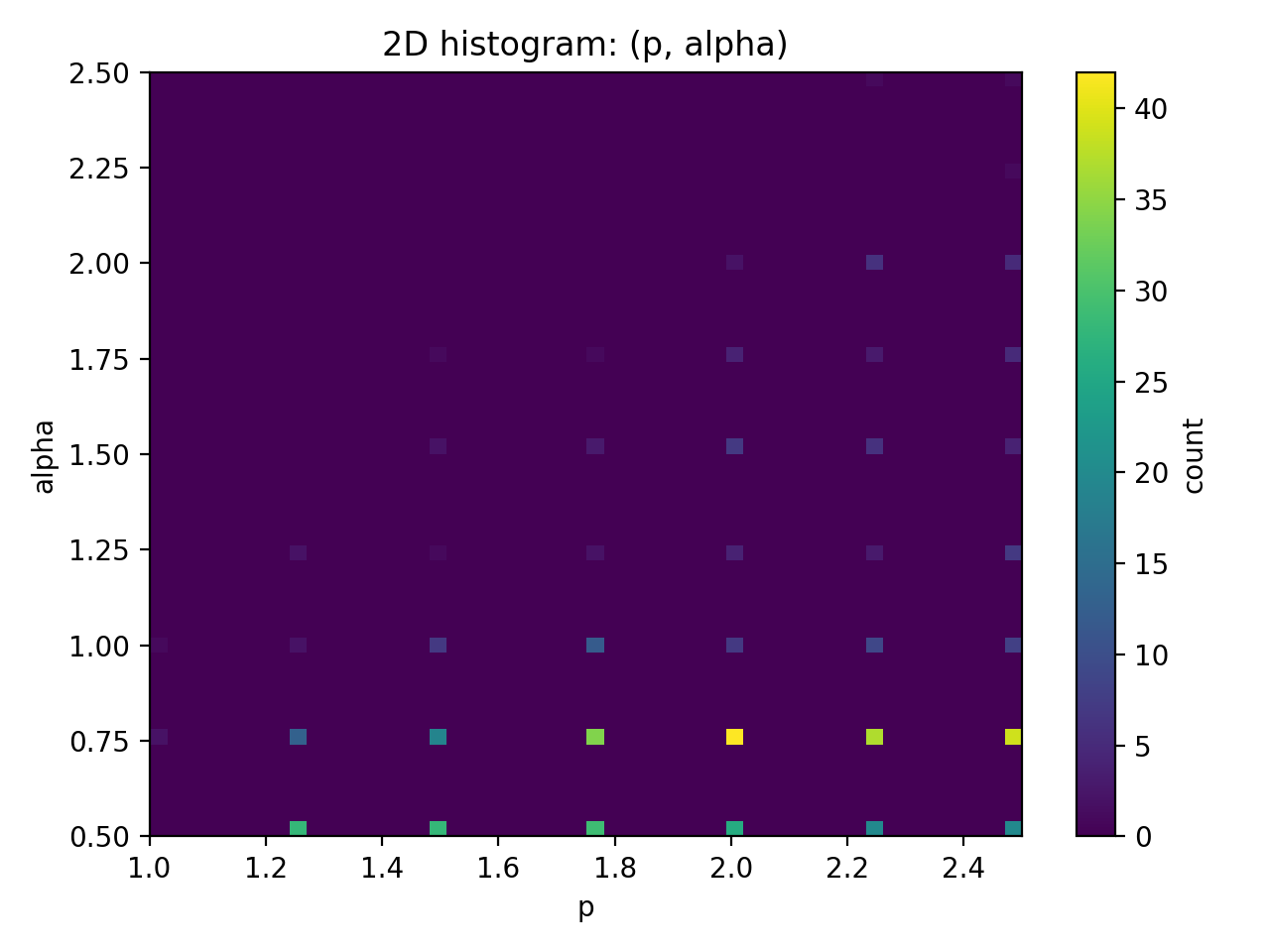}
         \caption{Top-5 $(p, \alpha)$}
         \label{fig:best_p_alpha}
    \end{subfigure}
    \caption{Histogram of top-5 hyperparameter selections $(\alpha, \lambda, p)$ for different OPT models with quantization bits of $q\in\{2,3,4,5\}$.}
    \label{fig:histogram}
\end{figure}

We evaluated the perplexity performance for OPT family when using uniformly sampled grids of $\alpha$, $\lambda$, and $p$.
We then selected 5 best combinations per OPT model and quantization bits $q$.
Then, we analyzed the histogram of those best parameter sets, as shown in Figure~\ref{fig:histogram}.
Those histogram plots may suggest some general trends:
\begin{itemize}
    \item The norm exponent $\alpha$ should be around $0.5$ or $0.75$. It suggests that the diagonal correlation approximation should be cooled down $\alpha < 1$ to be more uniform than amplified with $\alpha>1$.

    \item The damping factor $\lambda$ should be around $0.4$, which is way larger than the nominal choice like $\lambda=0.01$ in most literature. It may be because the dimension $d$ is very large compared to available token length $T$, leading to a larger shrinkage factor to be reliable under certain criterion~\cite{ledoit2004well}. Also, it indicates that selecting around $50\%$ damping factor offers a good trade-off between activation-aware loss and activation-unaware loss in (\ref{eq:lprime-app}).

    \item The best $\ell_p$-norm is around $p=2$, which can be well-justified from the derivation in (\ref{eq:lp2-app}). More importantly, $\ell_1$-norm was found to be a terrible choice. The original AWQ relied on a heuristic metric based on $\ell_1$, while some literature pointed out such an approximation may be suboptimal: e.g., LatentLLM~\cite{koike2025latentllm} made a thorough comparison to other variants of preconditioning; and Wanda~\cite{sun2023simple} derives a proper diagonal approximation with $\ell_2$ correlation.
    
\end{itemize}
In our experiments, the results of AWQ are based on $\ell_2$-norm not the original $\ell_1$-norm, while $\alpha$ is optimized with line search.

\section{Experiments Setup}
\label{sec:setup}

We conduct experiments for LLM benchmarks to evaluate the effectiveness of our method.
Our experiments are based on the same setting of SparseLLM~\cite{bai2024sparsellm} and their code base\footnote{\url{https://github.com/BaiTheBest/SparseLLM}}.
Following existing work~\cite{sun2023simple}, we quantize all linear layers in LLM transformers.
We implemented TTQ in PyTorch~\cite{paszke2019pytorch} and used the HuggingFace Transformers library \cite{wolf2019huggingface} for handling models and datasets. 
All experiments are conducted on NVIDIA A40 or A100 GPUs. 

For LLM experiments, we first consider the OPT model family~\cite{zhang2022opt} as it provides a wide range of model scales from 125M to 175B. 
We then evaluate more recent LLMs: Qwen3~\cite{yang2025qwen3} and Gemma3~\cite{team2025gemma}.
We show results on different sizes of models to provide a broader picture for the performance of TTQ. 
We measure perplexity score for three popular benchmarks: raw-WikiText2 (WT2)~\cite{merity2016pointer}; the Penn Treebank (PTB)~\cite{marcus1994penn}; and C4~\cite{raffel2020exploring}.
Details of LLMs and datasets we used are found in Appendices~\ref{sec:model} and \ref{sec:dataset}.

\section{Runtime Benchmark}
\label{sec:runtime}

We show TTQ can accelerate inference at test time over un-quantized LLMs, even with the extra overhead of prescaling and low-rank factors.
To evaluate the TTQ runtime speed, we use Marlin kernel~\cite{frantar2025marlin} which is one of best-known \texttt{int\_matmul} CUDA kernels.
Specifically, we use \texttt{vllm.\_custom\_ops.marlin\_gemm} in vllm library\footnote{\url{https://github.com/vllm-project/vllm}}.
It often outperforms \texttt{vllm.\_custom\_ops.awq\_gemm}.
We use CUDA 12.8, PyTorch 2.9.1, and vllm 0.15.1.
We include prescaling with $D$ diagonal matrix and low-rank projections with $B$ and $A$ on top of the Marlin kernel for TTQ like below:
\begin{lstlisting}
def forward(self, x): 
    # quantization params on the fly
    W, S, Z, D = self.find_params(x)
    # low-rank projection
    y0 = (x @ self.A.t().contiguous()) @ self.B.t().contiguous()
    # pre-scaling
    x = x / D
    # scaled int_matmul
    return marlin_gemm(x, y0, W, S, Z)

def find_params(self, x):
    # scale
    D = (x.norm(p=self.p, axis=0) + self.lam) ** self.alpha
    # prescale weight
    W = self.weight * D
    W = W.reshape(-1, self.groupsize)
    # scale, zeros
    Z = W.amin(axis=1, keepdim=True)
    S = (W.amax(axis=1, keepdim=True) - Z) / self.qmax
    # quantize
    W = ((W - Z) / S).round().clamp(0. self.qmax).to(torch.int32)    
    return W, S, Z, D
\end{lstlisting}
To reduce the overhead of multi-kernel launching, we use \texttt{torch.compile} and CUDA graph replay\footnote{\url{https://docs.vllm.ai/en/stable/design/cuda_graphs/}} for all cases.
We consider a decode phase of LLM projecting a single token.

Tables~\ref{tab:kernel_a40}, \ref{tab:kernel_a100}, \ref{tab:kernel_l40}, \ref{tab:kernel_rtx3090} and \ref{tab:kernel_rtx4090} show the runtime benchmark for Qwen3 models on NVIDIA A40, A100, L40, RTX3090 and RTX4090 GPUs, respectively.
To highlight the speedup, we focus on linear query projection module for benchmarking the original linear layer, offline 4-bit quanitzed AWQ, and online 4-bit quantized TTQ.
We can observe:
\begin{itemize}
    \item The throughput generally degrades with larger LLMs.
    \item AWQ can accelerate LLMs, especially with the Marlin kernel, up to 6.7 folds at 32B model on RTX4090.
    \item TTQ ($r=0$) has no significant loss in speed over AWQ, even with extra operations.
    \item Even without custom TTQ kernel in the presence of $r=16$ low-rank projections, TTQ can still accelerate the inference up to 4.9 folds at 32B model on RTX4090.
    \item TTQ shows more advantage on larger LLMs because of dominating weight traffic (e.g., Qwen3-32B model needs to transfer $5{,}120\times 8{,}192$ weights of 168MB from global memory to shared memory on GPUs for FP16 query projection).
\end{itemize}
AWQ fuses the prescaling $D$ into previous modules, while TTQ needs to fuse into \texttt{int\_matmul}. 
Most kernels are not yet compatible for such prologue fusion, and thus designing custom CUDA kernels for TTQ should further accelerate the inference on GPU.
In addition, most kernels are not designed for arbitrary bit width yet, and custom kernel design for 2-bit TTQ should speedup more, theoretically doubling due to traffic reduction with weight packing.

\begin{table*}[t]
\centering
\caption{Runtime Speed (k tokens/sec) of Query Projection Module for Qwen3 Models with 4-bit AWQ and TTQ, on NVIDIA A40 GPU.}
\label{tab:kernel_a40}
\small
\newcommand{\highscore}{80}
\setlength{\tabcolsep}{1pt}
\begin{tabular}{l rl rl rl rl rl rl rl}
\toprule
Qwen3 & 
\multicolumn{2}{c}{0.6B} &
\multicolumn{2}{c}{1.7B} &
\multicolumn{2}{c}{4B} &
\multicolumn{2}{c}{8B} &
\multicolumn{2}{c}{14B} &
\multicolumn{2}{c}{32B} 
\\
\midrule
FP16 & 
57.58 & \progressbar{57.58/\highscore}[0.9] &
41.29 & \progressbar{41.29/\highscore}[0.9] &
{21.38} & \progressbar{21.38/\highscore}[0.9] &
14.10 & \progressbar{14.10/\highscore}[0.9] &
10.43 & \progressbar{10.43/\highscore}[0.9] &
\tikzmarknode{s1}{6.86} & \progressbar{6.86/\highscore}[0.9] 
\\
AWQ (\texttt{awq\_gemm}) & 
\underline{75.76} & \progressbar{75.76/\highscore}[0.9] &
{71.60} & \progressbar{71.60/\highscore}[0.9] &
54.61 & \progressbar{54.61/\highscore}[0.9] &
32.68 & \progressbar{32.68/\highscore}[0.9] &
\underline{25.28} & \progressbar{25.28/\highscore}[0.9] &
18.28 & \progressbar{18.28/\highscore}[0.9] 
\\
AWQ (\texttt{marlin\_gemm}) & 
{73.32} & \progressbar{73.32/\highscore}[0.9] &
\underline{73.70} & \progressbar{73.70/\highscore}[0.9] &
\textbf{73.98} & \progressbar{73.98/\highscore}[0.9] &
\textbf{38.91} & \progressbar{38.91/\highscore}[0.9] &
\textbf{37.52} & \progressbar{36.52/\highscore}[0.9] &
\textbf{30.32} & \progressbar{30.32/\highscore}[0.9] 
\\
\rowcolor{lgreen}
\textbf{TTQ} ($r=0$) & 
\textbf{80.63} & \progressbar{80.63/\highscore}[0.9] &
\textbf{78.68} & \progressbar{78.68/\highscore}[0.9] &
\underline{69.11} & \progressbar{69.11/\highscore}[0.9] &
\underline{36.52} & \progressbar{36.52/\highscore}[0.9] &
{24.94} & \progressbar{24.94/\highscore}[0.9] &
\underline{19.34} & \progressbar{19.34/\highscore}[0.9] 
\\
\rowcolor{lgreen}
\textbf{TTQ} ($r=16$) & 
66.67 & \progressbar{66.67/\highscore}[0.9] &
59.08 & \progressbar{59.08/\highscore}[0.9] &
{49.67} & \progressbar{49.67/\highscore}[0.9] &
30.32 & \progressbar{30.32/\highscore}[0.9] &
23.78 & \progressbar{23.78/\highscore}[0.9] &
\tikzmarknode{e1}{17.43} & \progressbar{17.43/\highscore}[0.9] 
\\
\bottomrule
    \begin{tikzpicture}[remember picture, overlay, -{Stealth[scale=1.2]}, blue, thick]
        \node[draw, ellipse, fit=(e1), inner sep=0.5pt] (circ) {};
        \draw[->] (s1) to[out=-160, in=160, looseness=0.7] node[pos=0.25, above=0pt, rotate=60] {$\times 2.5$} (circ);
    \end{tikzpicture}
\end{tabular}
\end{table*}

\begin{table*}[t]
\centering
\caption{Runtime Speed (k tokens/sec) of Query Projection Module for Qwen3 Models with 4-bit AWQ and TTQ, on NVIDIA A100 GPU.}
\label{tab:kernel_a100}
\small
\newcommand{\highscore}{70}
\setlength{\tabcolsep}{1pt}
\begin{tabular}{l rl rl rl rl rl rl rl}
\toprule
Qwen3 & 
\multicolumn{2}{c}{0.6B} &
\multicolumn{2}{c}{1.7B} &
\multicolumn{2}{c}{4B} &
\multicolumn{2}{c}{8B} &
\multicolumn{2}{c}{14B} &
\multicolumn{2}{c}{32B} 
\\
\midrule
FP16 & 
\underline{68.43} & \progressbar{68.43/\highscore}[0.9] &
\textbf{68.91} & \progressbar{68.91/\highscore}[0.9] &
{42.76} & \progressbar{42.76/\highscore}[0.9] &
30.69 & \progressbar{30.69/\highscore}[0.9] &
20.16 & \progressbar{20.16/\highscore}[0.9] &
\tikzmarknode{s1}{15.90} & \progressbar{15.90/\highscore}[0.9] 
\\
AWQ (\texttt{awq\_gemm}) & 
{68.10} & \progressbar{68.10/\highscore}[0.9] &
{56.09} & \progressbar{56.09/\highscore}[0.9] &
48.20 & \progressbar{48.20/\highscore}[0.9] &
38.28 & \progressbar{38.28/\highscore}[0.9] &
{32.25} & \progressbar{32.25/\highscore}[0.9] &
24.35 & \progressbar{24.35/\highscore}[0.9] 
\\
AWQ (\texttt{marlin\_gemm}) & 
\textbf{69.41} & \progressbar{69.41/\highscore}[0.9] &
\underline{64.93} & \progressbar{64.93/\highscore}[0.9] &
\textbf{58.41} & \progressbar{58.41/\highscore}[0.9] &
\textbf{53.13} & \progressbar{53.13/\highscore}[0.9] &
\textbf{49.07} & \progressbar{49.07/\highscore}[0.9] &
\textbf{42.28} & \progressbar{42.28/\highscore}[0.9] 
\\
\rowcolor{lgreen}
\textbf{TTQ} ($r=0$) & 
{61.85} & \progressbar{61.85/\highscore}[0.9] &
{57.72} & \progressbar{57.72/\highscore}[0.9] &
\underline{52.67} & \progressbar{52.67/\highscore}[0.9] &
\underline{48.58} & \progressbar{48.58/\highscore}[0.9] &
\underline{45.17} & \progressbar{45.17/\highscore}[0.9] &
\underline{39.40} & \progressbar{39.40/\highscore}[0.9] 
\\
\rowcolor{lgreen}
\textbf{TTQ} ($r=16$) & 
46.04 & \progressbar{46.04/\highscore}[0.9] &
43.11 & \progressbar{43.11/\highscore}[0.9] &
{40.01} & \progressbar{40.01/\highscore}[0.9] &
37.19 & \progressbar{37.19/\highscore}[0.9] &
34.32 & \progressbar{34.32/\highscore}[0.9] &
\tikzmarknode{e1}{30.46} & \progressbar{30.46/\highscore}[0.9] 
\\
\bottomrule
    \begin{tikzpicture}[remember picture, overlay, -{Stealth[scale=1.2]}, blue, thick]
        \node[draw, ellipse, fit=(e1), inner sep=0.5pt] (circ) {};
        \draw[->] (s1) to[out=-160, in=160, looseness=0.7] node[pos=0.15, above=0pt, rotate=60] {$\times 1.9$} (circ);
    \end{tikzpicture}
\end{tabular}
\end{table*}

\begin{table*}[t]
\centering
\caption{Runtime Speed (k tokens/sec) of Query Projection Module for Qwen3 Models with 4-bit AWQ and TTQ, on NVIDIA L40 GPU.}
\label{tab:kernel_l40}
\small
\newcommand{\highscore}{95}
\setlength{\tabcolsep}{1pt}
\begin{tabular}{l rl rl rl rl rl rl rl}
\toprule
Qwen3 & 
\multicolumn{2}{c}{0.6B} &
\multicolumn{2}{c}{1.7B} &
\multicolumn{2}{c}{4B} &
\multicolumn{2}{c}{8B} &
\multicolumn{2}{c}{14B} &
\multicolumn{2}{c}{32B} 
\\
\midrule
FP16 & 
\textbf{91.35} & \progressbar{91.35/\highscore}[0.9] &
75.18 & \progressbar{75.18/\highscore}[0.9] &
{74.60} & \progressbar{74.60/\highscore}[0.9] &
53.31 & \progressbar{53.31/\highscore}[0.9] &
44.45 & \progressbar{44.45/\highscore}[0.9] &
\tikzmarknode{s1}{34.74} & \progressbar{34.74/\highscore}[0.9] 
\\
AWQ (\texttt{awq\_gemm}) & 
\underline{91.01} & \progressbar{91.01/\highscore}[0.9] &
{84.70} & \progressbar{84.70/\highscore}[0.9] &
76.00 & \progressbar{76.00/\highscore}[0.9] &
59.99 & \progressbar{59.99/\highscore}[0.9] &
{48.95} & \progressbar{48.95/\highscore}[0.9] &
47.87 & \progressbar{47.87/\highscore}[0.9] 
\\
AWQ (\texttt{marlin\_gemm}) & 
{86.57} & \progressbar{86.57/\highscore}[0.9] &
\underline{91.35} & \progressbar{91.35/\highscore}[0.9] &
\underline{91.28} & \progressbar{91.28/\highscore}[0.9] &
\textbf{90.93} & \progressbar{90.93/\highscore}[0.9] &
\textbf{77.81} & \progressbar{77.81/\highscore}[0.9] &
\textbf{68.15} & \progressbar{68.15/\highscore}[0.9] 
\\
\rowcolor{lgreen}
\textbf{TTQ} ($r=0$) & 
{90.59} & \progressbar{90.59/\highscore}[0.9] &
\textbf{92.57} & \progressbar{92.57/\highscore}[0.9] &
\textbf{91.86} & \progressbar{91.86/\highscore}[0.9] &
\underline{84.19} & \progressbar{84.19/\highscore}[0.9] &
\underline{71.28} & \progressbar{71.28/\highscore}[0.9] &
\underline{63.37} & \progressbar{63.37/\highscore}[0.9] 
\\
\rowcolor{lgreen}
\textbf{TTQ} ($r=16$) & 
72.82 & \progressbar{72.82/\highscore}[0.9] &
69.11 & \progressbar{69.11/\highscore}[0.9] &
{69.84} & \progressbar{69.84/\highscore}[0.9] &
63.37 & \progressbar{63.37/\highscore}[0.9] &
55.87 & \progressbar{55.87/\highscore}[0.9] &
\tikzmarknode{e1}{49.62} & \progressbar{49.00/\highscore}[0.9] 
\\
\bottomrule
    \begin{tikzpicture}[remember picture, overlay, -{Stealth[scale=1.2]}, blue, thick]
        \node[draw, ellipse, fit=(e1), inner sep=0.5pt] (circ) {};
        \draw[->] (s1) to[out=-160, in=160, looseness=0.5] node[pos=0.15, above=0pt, rotate=60] {$\times 1.4$} (circ);
    \end{tikzpicture}
\end{tabular}
\end{table*}

\begin{table*}[t]
\centering
\caption{Runtime Speed (k tokens/sec) of Query Projection Module for Qwen3 Models with 4-bit AWQ and TTQ, on NVIDIA RTX3090 GPU.}
\label{tab:kernel_rtx3090}
\small
\newcommand{\highscore}{70}
\setlength{\tabcolsep}{1pt}
\begin{tabular}{l rl rl rl rl rl rl rl}
\toprule
Qwen3 & 
\multicolumn{2}{c}{0.6B} &
\multicolumn{2}{c}{1.7B} &
\multicolumn{2}{c}{4B} &
\multicolumn{2}{c}{8B} &
\multicolumn{2}{c}{14B} &
\multicolumn{2}{c}{32B} 
\\
\midrule
FP16 & 
59.75 & \progressbar{59.75/\highscore}[0.9] &
54.38 & \progressbar{54.38/\highscore}[0.9] &
29.53 & \progressbar{29.53/\highscore}[0.9] &
20.37 & \progressbar{20.37/\highscore}[0.9] &
14.15 & \progressbar{14.15/\highscore}[0.9] &
\tikzmarknode{s1}{9.77} & \progressbar{9.77/\highscore}[0.9] 
\\
AWQ (\texttt{awq\_gemm}) & 
{62.89} & \progressbar{62.89/\highscore}[0.9] &
60.54 & \progressbar{60.54/\highscore}[0.9] &
48.39 & \progressbar{48.39/\highscore}[0.9] &
33.48 & \progressbar{33.48/\highscore}[0.9] &
27.92 & \progressbar{27.92/\highscore}[0.9] &
21.62 & \progressbar{21.62/\highscore}[0.9] 
\\
AWQ (\texttt{marlin\_gemm}) & 
\underline{63.13} & \progressbar{63.13/\highscore}[0.9] &
\textbf{67.86} & \progressbar{67.86/\highscore}[0.9] &
\textbf{65.88} & \progressbar{65.88/\highscore}[0.9] &
\textbf{46.97} & \progressbar{46.97/\highscore}[0.9] &
\textbf{36.30} & \progressbar{36.30/\highscore}[0.9] &
\textbf{26.17} & \progressbar{26.17/\highscore}[0.9] 
\\
\rowcolor{lgreen}
\textbf{TTQ} ($r=0$) & 
\textbf{68.07} & \progressbar{68.07/\highscore}[0.9] &
\underline{67.34} & \progressbar{67.34/\highscore}[0.9] &
\underline{64.43} & \progressbar{64.43/\highscore}[0.9] &
\underline{43.10} & \progressbar{43.10/\highscore}[0.9] &
\underline{34.25} & \progressbar{34.25/\highscore}[0.9] &
\underline{24.96} & \progressbar{24.96/\highscore}[0.9] 
\\
\rowcolor{lgreen}
\textbf{TTQ} ($r=16$) & 
57.57 & \progressbar{57.57/\highscore}[0.9] &
58.72 & \progressbar{58.72/\highscore}[0.9] &
44.73 & \progressbar{44.73/\highscore}[0.9] &
33.20 & \progressbar{33.20/\highscore}[0.9] &
28.45 & \progressbar{28.45/\highscore}[0.9] &
\tikzmarknode{e1}{21.85} & \progressbar{21.85/\highscore}[0.9] 
\\
\bottomrule
    \begin{tikzpicture}[remember picture, overlay, -{Stealth[scale=1.2]}, blue, thick]
        \node[draw, ellipse, fit=(e1), inner sep=0.5pt] (circ) {};
        \draw[->] (s1) to[out=-160, in=160, looseness=0.7] node[pos=0.25, above=0pt, rotate=60] {$\times 2.2$} (circ);
    \end{tikzpicture}
\end{tabular}
\end{table*}

\begin{table*}[t]
\centering
\caption{Runtime Speed (k tokens/sec) of Query Projection Module for Qwen3 Models with 4-bit AWQ and TTQ, on NVIDIA RTX4090 GPU.}
\label{tab:kernel_rtx4090}
\small
\newcommand{\highscore}{125}
\setlength{\tabcolsep}{1pt}
\begin{tabular}{l rl rl rl rl rl rl rl}
\toprule
Qwen3 & 
\multicolumn{2}{c}{0.6B} &
\multicolumn{2}{c}{1.7B} &
\multicolumn{2}{c}{4B} &
\multicolumn{2}{c}{8B} &
\multicolumn{2}{c}{14B} &
\multicolumn{2}{c}{32B} 
\\
\midrule
FP16 & 
\underline{116.82} & \progressbar{116.82/\highscore}[0.9] &
77.04 & \progressbar{77.04/\highscore}[0.9] &
72.45 & \progressbar{72.45/\highscore}[0.9] &
58.44 & \progressbar{58.44/\highscore}[0.9] &
46.78 & \progressbar{46.78/\highscore}[0.9] &
\tikzmarknode{s1}{10.76} & \progressbar{10.76/\highscore}[0.9] 
\\
AWQ (\texttt{awq\_gemm}) & 
{113.84} & \progressbar{113.84/\highscore}[0.9] &
90.42 & \progressbar{90.42/\highscore}[0.9] &
80.19 & \progressbar{80.19/\highscore}[0.9] &
62.94 & \progressbar{62.94/\highscore}[0.9] &
51.32 & \progressbar{51.32/\highscore}[0.9] &
50.45 & \progressbar{50.45/\highscore}[0.9] 
\\
AWQ (\texttt{marlin\_gemm}) & 
\textbf{120.62} & \progressbar{120.62/\highscore}[0.9] &
\textbf{115.20} & \progressbar{115.20/\highscore}[0.9] &
\textbf{112.61} & \progressbar{112.61/\highscore}[0.9] &
\textbf{101.00} & \progressbar{101.00/\highscore}[0.9] &
\textbf{80.45} & \progressbar{80.45/\highscore}[0.9] &
\textbf{72.34} & \progressbar{72.34/\highscore}[0.9] 
\\
\rowcolor{lgreen}
\textbf{TTQ} ($r=0$) & 
108.00 & \progressbar{108.00/\highscore}[0.9] &
\underline{104.11} & \progressbar{104.11/\highscore}[0.9] &
\underline{103.45} & \progressbar{103.45/\highscore}[0.9] &
\underline{92.23} & \progressbar{92.23/\highscore}[0.9] &
\underline{74.72} & \progressbar{74.72/\highscore}[0.9] &
\underline{67.40} & \progressbar{67.40/\highscore}[0.9] 
\\
\rowcolor{lgreen}
\textbf{TTQ} ($r=16$) & 
77.88 & \progressbar{77.88/\highscore}[0.9] &
73.47 & \progressbar{73.47/\highscore}[0.9] &
72.55 & \progressbar{72.55/\highscore}[0.9] &
66.62 & \progressbar{66.62/\highscore}[0.9] &
58.65 & \progressbar{58.65/\highscore}[0.9] &
\tikzmarknode{e1}{53.17} & \progressbar{53.17/\highscore}[0.9] 
\\
\bottomrule
    \begin{tikzpicture}[remember picture, overlay, -{Stealth[scale=1.2]}, blue, thick]
        \node[draw, ellipse, fit=(e1), inner sep=0.5pt] (circ) {};
        \draw[->] (s1) to[out=-160, in=160, looseness=0.7] node[pos=0.2, above=0pt, rotate=60] {$\times 4.9$} (circ);
    \end{tikzpicture}
\end{tabular}
\end{table*}

\section{LLM Benchmark Results}
\label{sec:llm}

Tables~\ref{tab:perp_opt_full}, \ref{tab:perp_qwen_full}, and \ref{tab:perp_gemma_full} show full performance results including standard deviation and more model variants for OPT, Qwen3, and Gemma3 LLMs, respectively. 

\begin{table*}[t]
\centering
\caption{Perplexity ($\downarrow$) of OPT models with different quantization methods. 
It shows macro average and standard deviation across WT2/PTB/C4 datasets.
Groupsize is $g=32$ for all cases. Calibration token length is $T=2^{17}$ for AWQ.
\textbf{Bold} and \underline{underline} denote the best and second best, respectively.
Asterisk ``*'' indicates reaching competitive performance to the original un-compressed LLM.
}
\label{tab:perp_opt_full}
\small
\begin{tabular}{l rrrr}
  \toprule
  Quantization $q$ & 
  2 bits & 3 bits & 4 bits & 5 bits  
  \\
  \midrule
  \multicolumn{5}{c}{OPT-125M (WT2: 27.7, PTB: 39.0, C4: 26.6, Avg: \textbf{31.1}$_{\pm6.8}$)}
  \\
  \midrule
  RTN & 
  5058.5$_{\pm 833.7}$  & 
  56.3$_{\pm 10.5}$  & 
  33.5$_{\pm 7.0}$  & 
  31.8$_{\pm 7.0}$  
  \\
  AWQ (WT2 Calib) & 
  381.7$_{\pm 159.8}$ &
  37.4$_{\pm 7.4}$ &
  32.3$_{\pm 6.5}$ &
  31.4$_{\pm 6.1}$ 
  \\
  AWQ (PTB Calib) &
  375.3$_{\pm 84.9}$ &
  37.3$_{\pm 6.9}$ &
  32.2$_{\pm 6.3}$ &
  31.3$_{\pm 6.0}$ 
  \\
  AWQ (C4 Calib) &
  451.7$_{\pm 181.7}$ &
  37.7$_{\pm 7.5}$ &
  32.6$_{\pm 6.6}$ &
  31.3$_{\pm 6.0}$ 
  \\
  \rowcolor{lgreen}
  \textbf{TTQ} ($r=0$) &
  \underline{257.4}$_{\pm 65.3}$ & 
  \underline{36.6}$_{\pm 8.2}$ &  
  \underline{31.9}$_{\pm 7.0}$ & 
  \underline{31.2}$_{\pm 6.8}$ 
  \\
  \rowcolor{lgreen}
  \textbf{TTQ} ($r=16$) &
  \textbf{141.7}$_{\pm 36.9}$ & 
  \textbf{35.8}$_{\pm 8.3}$ &  
  \textbf{31.8}$_{\pm 7.0}$ &  
  *\textbf{31.1}$_{\pm 6.8}$   
  \\
  \toprule
  \multicolumn{5}{c}{OPT-350M (WT2: 22.0, PTB: 31.1, C4: 22.6, Avg: \textbf{25.2}$_{\pm 5.1}$)}
  \\
  \midrule
  RTN & 
  25808.8$_{\pm 9601.0}$  & 
  55.3$_{\pm 13.8}$  & 
  27.6$_{\pm 5.5}$  & 
  25.6$_{\pm 5.3}$  
  \\
  AWQ (WT2 Calib) &
  \underline{293.2}$_{\pm 103.5}$ &
  28.8$_{\pm 5.1}$ &
  \underline{25.8}$_{\pm 4.5}$ &
  \underline{25.4}$_{\pm 4.4}$ 
  \\
  AWQ (PTB Calib) &
  350.7$_{\pm 107.0}$ &
  28.9$_{\pm 5.1}$ &
  \underline{25.8}$_{\pm 4.5}$ &
  \underline{25.4}$_{\pm 4.4}$ 
  \\
  AWQ (C4 Calib) &
  322.8$_{\pm 98.6}$ &
  29.0$_{\pm 5.6}$ &
  25.9$_{\pm 4.5}$ &
  \underline{25.4}$_{\pm 4.4}$ 
  \\
  \rowcolor{lgreen}
  \textbf{TTQ} ($r=0$) &
  {358.3}$_{\pm 123.9}$ & 
  \underline{28.6}$_{\pm 6.0}$ & 
  \textbf{25.7}$_{\pm 5.3}$ & 
  \textbf{25.3}$_{\pm 5.1}$ 
  \\
  \rowcolor{lgreen}
  \textbf{TTQ} ($r=16$) &
  \textbf{129.8}$_{\pm 46.8}$&
  \textbf{27.9}$_{\pm 5.9}$&
  \textbf{25.7}$_{\pm 5.2}$&
  \textbf{25.3}$_{\pm 5.1}$
  \\
  \toprule
  \multicolumn{5}{c}{OPT-1.3B (WT2: 14.6, PTB: 20.3, C4: 16.1, Avg: \textbf{17.0}$_{\pm 3.0}$)}
  \\
  \midrule
  RTN & 
  11514.4$_{\pm 2621.8}$  & 
  27.2$_{\pm 6.8}$  & 
  18.1$_{\pm 3.4}$  & 
  \underline{17.2}$_{\pm 3.1}$  
  \\
  AWQ (WT2 Calib) &
  32.4$_{\pm 8.6}$ &
  \underline{18.0}$_{\pm 2.8}$ &
  17.3$_{\pm 2.6}$ &
  *\textbf{17.0}$_{\pm 2.5}$ 
  \\
  AWQ (PTB Calib) &
  32.6$_{\pm 6.6}$ &
  18.1$_{\pm 2.7}$ &
  17.3$_{\pm 2.6}$ &
  *\textbf{17.0}$_{\pm 2.5}$ 
  \\
  AWQ (C4 Calib) &
  \underline{31.7}$_{\pm 8.1}$ &
  \underline{18.0}$_{\pm 2.8}$ &
  \underline{17.2}$_{\pm 2.6}$ &
  *\textbf{17.0}$_{\pm 2.5}$ 
  \\
  \rowcolor{lgreen}
  \textbf{TTQ} ($r=0$) &
  {32.2}$_{\pm 8.3}$ & 
  {18.2}$_{\pm 3.4}$ & 
  \underline{17.2}$_{\pm 3.0}$ & 
  *\textbf{17.0}$_{\pm 3.0}$ 
  \\
  \rowcolor{lgreen}
  \textbf{TTQ} ($r=16$) &
  \textbf{27.2}$_{\pm 6.1}$ & 
  \textbf{17.9}$_{\pm 3.1}$ & 
  \textbf{17.1}$_{\pm 3.0}$ & 
  *\textbf{17.0}$_{\pm 3.0}$ 
  \\
  \toprule
  \multicolumn{5}{c}{OPT-2.7B (WT2: 12.5, PTB: 18.0, C4: 14.3, Avg: \textbf{14.9}$_{\pm 2.8}$)}
  \\
  \midrule
  RTN & 
  6274.5$_{\pm 1350.7}$  & 
  36.0$_{\pm 11.5}$  & 
  15.7$_{\pm 3.1}$  & 
  \underline{15.0}$_{\pm 2.8}$  
  \\
  AWQ (WT2 Calib) & 
  23.1$_{\pm 5.3}$  & 
  \underline{15.7}$_{\pm 3.0}$  & 
  \underline{15.1}$_{\pm 2.8}$  & 
  \underline{15.0}$_{\pm 2.8}$ 
  \\
  AWQ (PTB Calib) & 
  23.2$_{\pm 4.0}$  & 
  \underline{15.7}$_{\pm 3.0}$  & 
  \textbf{15.0}$_{\pm 2.9}$  & 
  \underline{15.0}$_{\pm 2.9}$   
  \\
  AWQ (C4 Calib) & 
  \underline{22.9}$_{\pm 5.1}$  & 
  \underline{15.7}$_{\pm 3.0}$  & 
  \textbf{15.0}$_{\pm 2.8}$  & 
  \underline{15.0}$_{\pm 2.8}$   
  \\
  \rowcolor{lgreen}
  \textbf{TTQ} ($r=0$) &
  {23.7}$_{\pm 5.1}$ & 
  \underline{15.7}$_{\pm 3.0}$ &  
  \textbf{15.0}$_{\pm 2.8}$ & 
  *\textbf{14.9}$_{\pm 2.8}$ 
  \\
  \rowcolor{lgreen}
  \textbf{TTQ} ($r=16$) &
  \textbf{21.2}$_{\pm 4.5}$ & 
  \textbf{15.5}$_{\pm 2.9}$ &  
  \textbf{15.0}$_{\pm 2.8}$ & 
  *\textbf{14.9}$_{\pm 2.8}$ 
  \\
  \toprule
  \multicolumn{5}{c}{OPT-6.7B (WT2: 10.9, PTB: 15.8, C4: 12.7, Avg: \textbf{13.1}$_{\pm 2.5}$)}
  \\
  \midrule
  RTN & 
  5716.5$_{\pm 664.0}$  & 
  26.2$_{\pm 8.9}$  & 
  13.7$_{\pm 2.8}$  & 
  \underline{13.2}$_{\pm 2.5}$  
  \\
  AWQ (WT2 Calib) &
  17.2$_{\pm 3.6}$ &
  \underline{13.5}$_{\pm 2.6}$ &
  \underline{13.2}$_{\pm 2.5}$ &
  *\textbf{13.1}$_{\pm 2.5}$ 
  \\
  AWQ (PTB Calib) &
  17.1$_{\pm 3.0}$ &
  13.6$_{\pm 2.6}$ &
  \underline{13.2}$_{\pm 2.5}$ &
  *\textbf{13.1}$_{\pm 2.5}$ 
  \\
  AWQ (C4 Calib) &
  \underline{16.9}$_{\pm 3.5}$ &
  13.6$_{\pm 2.6}$ &
  \underline{13.2}$_{\pm 2.6}$ &
  *\textbf{13.1}$_{\pm 2.5}$ 
  \\
  \rowcolor{lgreen}
  \textbf{TTQ} ($r=0$) &
  {17.2}$_{\pm 3.5}$ & 
  {13.6}$_{\pm 2.6}$ &  
  \underline{13.2}$_{\pm 2.5}$ &
  *\textbf{13.1}$_{\pm 2.5}$ 
  \\
  \rowcolor{lgreen}
  \textbf{TTQ} ($r=16)$&
  \textbf{16.3}$_{\pm 3.4}$ & 
  \textbf{13.4}$_{\pm 2.6}$ &  
  *\textbf{13.1}$_{\pm 2.5}$ &
  *\textbf{13.1}$_{\pm 2.5}$ 
  \\
  \toprule
  \multicolumn{5}{c}{OPT-13B (WT2: 10.1, PTB: 14.5, C4: 12.1, Avg: \textbf{12.2}$_{\pm 2.2}$)}
  \\
  \midrule
  RTN & 
  6413.1$_{\pm 1051.0}$  & 
  15.5$_{\pm 3.1}$  & 
  \underline{12.5}$_{\pm 2.4}$  & 
  \underline{12.3}$_{\pm 2.3}$  
  \\
  AWQ (WT2 Calib) &
  15.3$_{\pm 2.9}$ &
  \underline{12.6}$_{\pm 2.3}$ &
  \textbf{12.3}$_{\pm 2.2}$ &
  *\textbf{12.2}$_{\pm 2.2}$ 
  \\
  AWQ (PTB Calib) &
  15.4$_{\pm 2.7}$ &
  \underline{12.6}$_{\pm 2.3}$ &
  \textbf{12.3}$_{\pm 2.2}$ &
  *\textbf{12.2}$_{\pm 2.2}$ 
  \\
  AWQ (C4 Calib) &
  15.2$_{\pm 2.9}$ &
  \underline{12.6}$_{\pm 2.3}$ &
  \textbf{12.3}$_{\pm 2.2}$ &
  *\textbf{12.2}$_{\pm 2.2}$ 
  \\
  \rowcolor{lgreen}
  \textbf{TTQ} ($r=0$) &
  \underline{15.0}$_{\pm 2.9}$ & 
  \textbf{12.5}$_{\pm 2.7}$ &  
  \textbf{12.3}$_{\pm 2.2}$ & 
  *\textbf{12.2}$_{\pm 2.2}$  
  \\
  \rowcolor{lgreen}
  \textbf{TTQ} ($r=16$) &
  \textbf{14.8}$_{\pm 2.8}$ & 
  \textbf{12.5}$_{\pm 2.2}$ & 
  \textbf{12.3}$_{\pm 2.2}$ & 
  *\textbf{12.2}$_{\pm 2.2}$ 
  \\
  \bottomrule
\end{tabular}
\end{table*}

\begin{table*}
\centering
\caption{Perplexity ($\downarrow$) of Qwen3 models with different quantization methods. 
It shows macro average and standard deviation across WT2/PTB/C4 datasets.
Groupsize is $g=32$ for all cases. Calibration token length is $T=2^{17}$ for AWQ.
\textbf{Bold} and \underline{underline} denote the best and second best, respectively.
Asterisk ``*'' indicates reaching competitive performance to the original un-compressed LLM.
}
\label{tab:perp_qwen_full}
\small
\begin{tabular}{l rrrr}
  \toprule
  Quantization $q$ & 
  2 bits & 3 bits & 4 bits & 5 bits  
  \\
  \midrule
  \multicolumn{5}{c}{Qwen3-0.6B (WT2: 21.0, PTB: 43.8, C4: 30.3, Avg:  \textbf{31.7}$_{\pm 11.5}$)}
  \\
  \midrule
  RTN & 
  8.2e6$_{\pm 4.4e6}$  & 
  127.3$_{\pm 54.1}$  & 
  38.2$_{\pm 14.7}$  & 
  33.5$_{\pm 12.4}$  
  \\
  AWQ (WT2 Calib) & 
  9739.1$_{\pm 2765.7}$ &
  49.4$_{\pm 20.2}$ &
  33.5$_{\pm 11.7}$ &
  32.4$_{\pm 11.8}$ 
  \\
  AWQ (PTB Calib) &
  17344.3$_{\pm 1962.7}$ &
  47.9$_{\pm 17.3}$ &
  34.1$_{\pm 12.5}$ &
  \underline{32.0}$_{\pm 11.5}$ 
  \\
  AWQ (C4 Calib) &
  5388.1$_{\pm 1862.6}$ &
  48.3$_{\pm 17.3}$ &
  \underline{33.0}$_{\pm 11.6}$ &
  32.1$_{\pm 11.8}$ 
  \\
  \rowcolor{lgreen}
  \textbf{TTQ} ($r=0$) &
  \underline{2827.8}$_{\pm 883.8}$ & 
  \underline{44.7}$_{\pm 16.9}$ &  
  \underline{33.0}$_{\pm 12.1}$ & 
  \textbf{31.9}$_{\pm 11.3}$ 
  \\
  \rowcolor{lgreen}
  \textbf{TTQ} ($r=16$) &
  \textbf{1552.6}$_{\pm 386.3}$ & 
  \textbf{42.0}$_{\pm 16.0}$ &  
  \textbf{32.9}$_{\pm 12.2}$ &  
  \textbf{31.9}$_{\pm 11.3}$   
  \\
  \toprule
  \multicolumn{5}{c}{Qwen3-1.7B (WT2: 16.7, PTB: 33.8, C4: 16.1, Avg: \textbf{24.2}$_{\pm 8.7}$)}
  \\
  \midrule
  RTN & 
  1.4e6$_{\pm 2.2e5}$  & 
  162.8$_{\pm 107.4}$  & 
  30.6$_{\pm 12.2}$  & 
  26.1$_{\pm 9.6}$  
  \\
  AWQ (WT2 Calib) &
  {1864.7}$_{\pm 2051.9}$ &
  30.0$_{\pm 12.0}$ &
  {24.5}$_{\pm 8.7}$ &
  {24.4}$_{\pm 8.6}$ 
  \\
  AWQ (PTB Calib) &
  2309.6$_{\pm 2524.0}$ &
  29.8$_{\pm 11.6}$ &
  {24.7}$_{\pm 8.3}$ &
  {24.5}$_{\pm 8.9}$ 
  \\
  AWQ (C4 Calib) &
  2364.7$_{\pm 2878.7}$ &
  28.2$_{\pm 9.9}$ &
  24.9$_{\pm 9.1}$ &
  {24.4}$_{\pm 8.8}$ 
  \\
  \rowcolor{lgreen}
  \textbf{TTQ} ($r=0$) &
  \underline{522.6}$_{\pm 259.4}$ & 
  \underline{27.3}$_{\pm 9.6}$  &
  \underline{24.4}$_{\pm 8.5}$    
  &
  \underline{24.3}$_{\pm 8.5}$    
  \\
  \rowcolor{lgreen}
  \textbf{TTQ} ($r=16$) &
  \textbf{264.6}$_{\pm 90.4}$ & 
  \textbf{26.4}$_{\pm 9.3}$ & 
  \textbf{24.3}$_{\pm 8.8}$ &   
  *\textbf{24.1}$_{\pm 8.5}$
  \\
  \toprule
  \multicolumn{5}{c}{Qwen3-4B (WT2: 13.7, PTB: 24.8, C4: 19.9, Avg: \textbf{19.4}$_{\pm 5.6}$)}
  \\
  \midrule
  RTN & 
  5803.3$_{\pm 2179.1}$  & 
  28.6$_{\pm 10.1}$  & 
  21.4$_{\pm 6.4}$  & 
  \underline{19.6}$_{\pm 5.5}$  
  \\
  AWQ (WT2 Calib) &
  {180.8}$_{\pm 65.3}$ &
  21.2$_{\pm 6.4}$ &
  {20.1}$_{\pm 5.7}$ &
  *\textbf{19.4}$_{\pm 5.6}$ 
  \\
  AWQ (PTB Calib) &
  199.2$_{\pm 29.7}$ &
  26.6$_{\pm 9.0}$ &
  \underline{19.8}$_{\pm 5.9}$ &
  *\textbf{19.4}$_{\pm 5.6}$ 
  \\
  AWQ (C4 Calib) &
  246.9$_{\pm 85.2}$ &
  21.7$_{\pm 6.3}$ &
  19.9$_{\pm 5.5}$ &
  \underline{19.6}$_{\pm 5.6}$ 
  \\
  \rowcolor{lgreen}
  \textbf{TTQ} ($r=0$) &
  \underline{120.4}$_{\pm 34.3}$ & 
  \textbf{20.9}$_{\pm 6.0}$  &
  \textbf{19.6}$_{\pm 5.5}$    
  &
  *\textbf{19.4}$_{\pm 5.5}$    
  \\
  \rowcolor{lgreen}
  \textbf{TTQ} ($r=16$) &
  \textbf{78.1}$_{\pm 19.5}$ & 
  \underline{21.1}$_{\pm 6.2}$  &
  \textbf{19.6}$_{\pm 5.6}$    
  &
  *\textbf{19.4}$_{\pm 5.5}$    
  \\
  \toprule
  \multicolumn{5}{c}{Qwen3-8B (WT2: 9.7, PTB: 17.2, C4: 15.4, Avg: \textbf{14.1}$_{\pm 3.9}$)}
  \\
  \midrule
  RTN & 
  5366.8$_{\pm 1985.5}$  & 
  18.4$_{\pm 5.7}$  & 
  14.9$_{\pm 4.1}$  & 
  14.4$_{\pm 3.9}$  
  \\
  AWQ (WT2 Calib) &
  {39.5}$_{\pm 12.9}$ &
  15.3$_{\pm 4.2}$ &
  {14.4}$_{\pm 4.0}$ &
  \underline{14.2}$_{\pm 4.0}$ 
  \\
  AWQ (PTB Calib) &
  39.4$_{\pm 9.3}$ &
  15.4$_{\pm 4.2}$ &
  {14.4}$_{\pm 4.0}$ &
  \underline{14.2}$_{\pm 3.9}$ 
  \\
  AWQ (C4 Calib) &
  37.6$_{\pm 10.9}$ &
  \underline{15.2}$_{\pm 4.1}$ &
  {14.4}$_{\pm 4.0}$ &
  *\textbf{14.1}$_{\pm 3.9}$ 
  \\
  \rowcolor{lgreen}
  \textbf{TTQ} ($r=0$) &
  \underline{32.3}$_{\pm 8.8}$ & 
  \textbf{15.0}$_{\pm 4.1}$ & 
  \textbf{14.2}$_{\pm 3.9}$ & 
  *\textbf{14.1}$_{\pm 3.9}$  
  \\
  \rowcolor{lgreen}
  \textbf{TTQ} ($r=16$) &
  \textbf{30.2}$_{\pm 8.0}$ & 
  \textbf{15.0}$_{\pm 4.1}$ & 
  \underline{14.3}$_{\pm 3.9}$ & 
  *\textbf{14.1}$_{\pm 3.9}$  
  \\
  \toprule
  \multicolumn{5}{c}{Qwen3-14B (WT2: 8.6, PTB: 15.2, C4: 13.8, Avg: \textbf{12.6}$_{\pm 3.5}$)}
  \\
  \midrule
  RTN & 
  366.3$_{\pm 107.7}$  & 
  15.6$_{\pm 4.6}$  & 
  12.9$_{\pm 3.7}$  & 
  \underline{12.8}$_{\pm 3.6}$  
  \\
  AWQ (WT2 Calib) &
  {23.0}$_{\pm 7.4}$ &
  13.4$_{\pm 3.7}$ &
  \underline{12.7}$_{\pm 3.5}$ &
  *\textbf{12.6}$_{\pm 3.4}$ 
  \\
  AWQ (PTB Calib) &
  23.0$_{\pm 6.7}$ &
  13.3$_{\pm 3.7}$ &
  \underline{12.7}$_{\pm 3.5}$ &
  *\textbf{12.6}$_{\pm 3.5}$ 
  \\
  AWQ (C4 Calib) &
  23.3$_{\pm 7.5}$ &
  13.4$_{\pm 3.7}$ &
  \underline{12.7}$_{\pm 3.5}$ &
  *\textbf{12.6}$_{\pm 3.4}$ 
  \\
  \rowcolor{lgreen}
  \textbf{TTQ} ($r=0$) &
  \underline{21.1}$_{\pm 6.5}$ & 
  \underline{13.2}$_{\pm 3.6}$ & 
  \underline{12.7}$_{\pm 3.5}$ & 
  *\textbf{12.6}$_{\pm 3.5}$  
  \\
  \rowcolor{lgreen}
  \textbf{TTQ} ($r=16$) &
  \textbf{20.3}$_{\pm 6.4}$ & 
  \textbf{13.1}$_{\pm 3.6}$ & 
  *\textbf{12.6}$_{\pm 3.5}$ & 
  *\textbf{12.6}$_{\pm 3.5}$  
  \\
  \bottomrule
\end{tabular}
\end{table*}

\begin{table*}
\centering
\caption{Perplexity ($\downarrow$) of Gemma3 models with different quantization methods. 
It shows macro average and standard deviation across WT2/PTB/C4 datasets.
Groupsize is $g=32$ for all cases. Calibration token length is $T=2^{17}$ for AWQ.
\textbf{Bold} and \underline{underline} denote the best and second best, respectively.
Asterisk ``*'' indicates reaching competitive performance to the original un-compressed LLM.
}
\label{tab:perp_gemma_full}
\small
\begin{tabular}{l rrrr}
  \toprule
  Quantization $q$ & 
  2 bits & 3 bits & 4 bits & 5 bits  
  \\
  \midrule
  \multicolumn{5}{c}{Gemma3-270M (WT2: 70.1, PTB: 698.7, C4: 68.1, Avg: \textbf{279.0}$_{\pm 353.5}$)}
  \\
  \midrule
  RTN & 
  2.6e11$_{\pm 1.2e11}$  & 
  1795.0$_{\pm 2207.5}$  & 
  391.9$_{\pm 505.5}$  & 
  315.0$_{\pm 410.9}$  
  \\
  AWQ (WT2 Calib) & 
  89188.4$_{\pm 58307.2}$ &
  517.4$_{\pm 690.0}$ &
  279.6$_{\pm 348.6}$ &
  306.0$_{\pm 405.2}$ 
  \\
  AWQ (PTB Calib) &
  1.9e5$_{\pm 2.0e9}$ &
  537.7$_{\pm 711.8}$ &
  310.0$_{\pm 408.2}$ &
  293.4$_{\pm 386.4}$ 
  \\
  AWQ (C4 Calib) &
  1.3e5$_{\pm 1.6e5}$ &
  598.1$_{\pm 832.9}$ &
  326.4$_{\pm 420.5}$ &
  *276.8$_{\pm 358.0}$ 
  \\
  \rowcolor{lgreen}
  \textbf{TTQ} ($r=0$) &
  \underline{30199.0}$_{\pm 23705.9}$ & 
  \underline{387.7}$_{\pm 489.0}$ &  
  *\underline{277.0}$_{\pm 357.1}$ & 
  *\underline{262.5}$_{\pm 335.2}$ 
  \\
  \rowcolor{lgreen}
  \textbf{TTQ} ($r=16$) &
  \textbf{19657.0}$_{\pm 17934.2}$ &
  \textbf{382.8}$_{\pm 486.2}$ &
  *\textbf{275.7}$_{\pm 353.6}$ &
  *\textbf{261.5}$_{\pm 343.1}$ 
  \\
  \toprule
  \multicolumn{5}{c}{Gemma3-1B (WT2: 27.7, PTB: 212.4, C4: 33.2, Avg: \textbf{91.1}$_{\pm 105.1}$)}
  \\
  \midrule
  RTN & 
  8.6e5$_{\pm 7.7e5}$  & 
  209.4$_{\pm 253.2}$  & 
  111.1$_{\pm 130.0}$  & 
  96.6$_{\pm 112.0}$  
  \\
  AWQ (WT2 Calib) &
  {4734.7}$_{\pm 5394.7}$ &
  134.7$_{\pm 159.3}$ &
  {91.9}$_{\pm 104.1}$ &
  {95.9}$_{\pm 112.2}$ 
  \\
  AWQ (PTB Calib) &
  9326.6$_{\pm 11932.8}$ &
  138.9$_{\pm 167.5}$ &
  {99.2}$_{\pm 115.1}$ &
  {95.5}$_{\pm 112.0}$ 
  \\
  AWQ (C4 Calib) &
  5486.9$_{\pm 6662.0}$ &
  150.8$_{\pm 186.3}$ &
  93.5$_{\pm 106.0}$ &
  {93.6}$_{\pm 108.4}$ 
  \\
  \rowcolor{lgreen}
  \textbf{TTQ} ($r=0$) &
  \underline{1928.5}$_{\pm 2313.2}$ &
  \underline{127.3}$_{\pm 150.5}$ &
  *\textbf{89.9}$_{\pm 100.8}$ &
  *\textbf{90.2}$_{\pm 103.3}$ 
  \\
  \rowcolor{lgreen}
  \textbf{TTQ} ($r=16$) &
  \textbf{1804.9}$_{\pm 1850.3}$ &
  \textbf{114.5}$_{\pm 130.8}$ &
  \underline{91.7}$_{\pm 107.4}$ &
  *\underline{90.3}$_{\pm 103.6}$ 
  \\
  \toprule
  \multicolumn{5}{c}{Gemma3-4B (WT2: 17.4, PTB: 641.5, C4: 23.3, Avg: \textbf{227.4}$_{\pm 358.6}$)}
  \\
  \midrule
  RTN & 
  8.1e6$_{\pm 1.4e7}$  & 
  606.4$_{\pm 997.2}$  & 
  251.0$_{\pm 395.1}$  & 
  257.8$_{\pm 410.5}$  
  \\
  AWQ (WT2 Calib) &
  5667.3$_{\pm 9506.0}$  & 
  351.7$_{\pm 565.7}$  & 
  *225.3$_{\pm 353.5}$  & 
  *\underline{221.9}$_{\pm 348.7}$   
  \\
  AWQ (PTB Calib) &
  4444.0$_{\pm 7433.4}$  & 
  352.9$_{\pm 571.5}$  & 
  240.4$_{\pm 380.6}$  & 
  240.5$_{\pm 380.9}$   
  \\
  AWQ (C4 Calib) &
  4938.5$_{\pm 8259.2}$  & 
  366.3$_{\pm 593.1}$  & 
  245.6$_{\pm 388.9}$  & 
  245.8$_{\pm 389.7}$   
  \\
  \rowcolor{lgreen}
  \textbf{TTQ} ($r=0$) &
  \textbf{3561.6}$_{\pm 6003.4}$  & 
  \textbf{239.5}$_{\pm 375.8}$  & 
  *\underline{222.2}$_{\pm 349.4}$  & 
  *\textbf{221.7}$_{\pm 349.1}$    \\
  \rowcolor{lgreen}
  \textbf{TTQ} ($r=16$) &
  \underline{3974.9}$_{\pm 6696.8}$  & 
  \underline{251.7}$_{\pm 396.6}$  & 
  *\textbf{222.0}$_{\pm 349.1}$  & 
  *\underline{221.9}$_{\pm 353.6}$  
  \\
  \toprule
  \multicolumn{5}{c}{Gemma3-12B (WT2: 25.1, PTB: 283.8, C4: 35.0, Avg: \textbf{114.6}$_{\pm 146.6}$)}
  \\
  \midrule
  RTN & 
  15467.0$_{\pm 13512.6}$  & 
  294.9$_{\pm 390.2}$  & 
  146.6$_{\pm 196.1}$  & 
  *108.3$_{\pm 136.5}$  
  \\
  AWQ (WT2 Calib) &
  1700.6$_{\pm 2328.9}$  & 
  147.4$_{\pm 196.4}$  & 
  125.9$_{\pm 166.1}$  & 
  114.7$_{\pm 149.9}$  
  \\
  AWQ (PTB Calib) &
  4193.8$_{\pm 6142.5}$  & 
  128.3$_{\pm 164.6}$  & 
  117.3$_{\pm 146.2}$  & 
  122.3$_{\pm 157.6}$  
  \\
  AWQ (C4 Calib) &
  \underline{1464.4}$_{\pm 2003.8}$  & 
  *100.8$_{\pm 122.2}$  & 
  146.0$_{\pm 197.0}$  & 
  132.0$_{\pm 173.5}$  
  \\
  \rowcolor{lgreen}
  \textbf{TTQ} ($r=0$) &
  \textbf{1295.6}$_{\pm 1743.4}$  & 
  *\textbf{84.6}$_{\pm 94.1}$  & 
  *\textbf{86.6}$_{\pm 102.3}$  & 
  *\textbf{92.9}$_{\pm 112.4}$  
  \\
  \rowcolor{lgreen}
  \textbf{TTQ} ($r=16$) &
  {1620.9}$_{\pm 2248.4}$  & 
  *\underline{100.7}$_{\pm 116.9}$  & 
  *\underline{91.2}$_{\pm 109.4}$  & 
  *\underline{98.3}$_{\pm 121.1}$  
  \\
  \bottomrule
\end{tabular}
\end{table*}

\section{VLM Benchmark Results}
\label{sec:vlm}

We show the quantization results for Qwen3-VL VLM models on TextVQA benchmark.
We quantize all linear modules except LM head and vision encoder.
Table~\ref{tab:vlm} shows VQA soft accuracy performance with different quantization methods.
For AWQ, we use different calibration dataset: COCO-Caption; OK-VQA; ChartQA; and TextVQA.
We confirm that our TTQ achieves best overall performance.
It is interesting to see that some quantized models can slightly outperform un-quantized Qwen-VL models.

\begin{table}[t]
    \centering
    \caption{Accuracy ($\uparrow$) of Qwen3-VL models with different quantization methods on TextVQA benchmark. 
    Groupsize is g = 32 for all cases.
    \textbf{Bold} and \underline{underline} denote the best and second best, respectively.
    Asterisk ``*'' indicates reaching competitive performance to the original un-compressed VLM.
    }
    \label{tab:vlm}
    \begin{tabular}{lrrrr}
    \toprule
    Quantization $q$ & 2 bits
    & 3 bits & 4 bits & 5 bits \\
    \midrule
    \multicolumn{5}{c}{Qwen3-VL-2B: Acc 80.35\%}
    \\
    \midrule
    RTN & 0.00\% & 9.23\% & 73.27\% & {80.22}\%
    \\
    AWQ (COCO-Cap Calib) & 1.41\% & 75.67\% & 79.04\% & 79.89\% 
    \\
    AWQ (OK-VQA Calib) & 1.37\% & {74.68}\% & {79.47}\% & 79.80\% 
    \\
    AWQ (ChartQA Calib) & 0.67\% & {75.52}\% & {79.29}\% & 79.80\% 
    \\
    AWQ (TextVQA Calib) & 0.82\% & {72.15}\% & {78.85}\% & 80.21\% 
    \\
    \rowcolor{lgreen}
    \textbf{TTQ} ($r=0$) &
    \underline{1.42}\% & \textbf{76.59}\% & \underline{79.59}\%  & \underline{80.34}\%
    \\
    \rowcolor{lgreen}
    \textbf{TTQ} ($r=16$) &
    \textbf{1.93}\% & \underline{76.01}\% & \textbf{79.67}\% & *\textbf{80.38}\%
    \\
    \toprule
    \multicolumn{5}{c}{Qwen3-VL-4B: Acc 81.44\%}
    \\
    \midrule
    RTN & 0.03\% & 74.79\% & 80.77\% & *81.47\% 
    \\
    AWQ (COCO-Cap Calib) & 0.13\% & 78.68\% & 80.49\% & 81.20\% 
    \\
    AWQ (OK-VQA Calib) & 0.18\% & 77.89\% & 80.58\% & 80.75\% 
    \\
    AWQ (ChartQA Calib) & 0.42\% & 77.37\% & 80.67\% & *81.47\% 
    \\
    AWQ (TextVQA Calib) & 0.25\% & {77.49}\% & {80.28}\% & 81.01\%
    \\
    \rowcolor{lgreen}
    \textbf{TTQ} ($r=0$) &
    \underline{1.51}\% & \textbf{79.93}\% & \textbf{81.29}\%  & *\underline{81.48}\%
    \\
    \rowcolor{lgreen}
    \textbf{TTQ} ($r=16$) &
    \textbf{7.47}\% & \underline{79.45}\% & \underline{81.22}\% & *\textbf{81.49}\%
    \\
    \toprule
    \multicolumn{5}{c}{Qwen3-VL-8B: Acc 81.72\%}
    \\
    \midrule
    RTN & 0.17\% & 78.51\% & 80.63\% & *81.73\%
    \\
    AWQ (COCO-Cap Calib) & 41.39\% & 80.37\% & 81.52\% & *81.79\% 
    \\
    AWQ (OK-VQA Calib) & 31.62\% & 79.51\% & 81.01\% & 81.55\% 
    \\
    AWQ (ChartQA Calib) & 41.17\% & 78.65\% & 81.23\% & 81.39\% 
    \\
    AWQ (TextVQA Calib) & 39.60\% & {78.43}\% & {81.25}\% & 81.69\% 
    \\
    \rowcolor{lgreen}
    \textbf{TTQ} ($r=0$) &
    \underline{42.20}\% & \underline{80.77}\% & \underline{81.57}\% & *\underline{81.81}\%
    \\
    \rowcolor{lgreen}
    \textbf{TTQ} ($r=16$) &
    \textbf{47.22}\% & \textbf{81.04}\% & \textbf{81.71}\% & *\textbf{81.85}\%
    \\
    \bottomrule
    \end{tabular}
\end{table}

\section{VLA Benchmark Results}
\label{sec:vla}

We show the quantization results for $\pi_{0.5}$ VLA model on LIBERO robot manipulation benchmark in Table~\ref{tab:vla}.
We quantize all linear modules for VLM backbone of $\pi_{0.5}$ model except LM head and vision encoder.
Here, we use 200 episodes per benchmark.
For AWQ, we compare with different calibration dataset from LIBERO Spatial, Object, Goal, and 10 task suites.
Although AWQ performs much better than RTN, its performance highly depends on offline calibration data. 
TTQ achieves best in average success rates.
The advantage is clearer at the long-horizon tasks of LIBERO 10.
It is interesting to see that our TTQ can allow 2-bit quantization for $\pi_{0.5}$  without causing much performance loss.

\begin{table}[t]
    \centering
    \caption{Success rate ($\uparrow$) of $\pi_{0.5}$ VLA model with different quantization methods on LIBERO robot manipulation benchmark. 
    We use $q=2$ bits and $g=64$ groupsize.
    \textbf{Bold} and \underline{underline} denote the best and second best, respectively.
    }
    \label{tab:vla}
    \newcolumntype{g}{>{\columncolor{lblue}}r}
    \begin{tabular}{lrrrrg}
    \toprule
    Benthmark & 
    Libero Spatial & Libero Object & Libero Goal & Libero 10 & Avg \\
    \midrule
    BF16 &
    97.5\% & 100.0\% & 97.0\% & 96.5\% & 97.75\% 
    \\
    \addlinespace
    RTN & 57.0\% & 65.0\% & 27.5\% & 2.0\% & 37.88\%
    \\
    AWQ (Spatial Calib) & 90.5\% & \textbf{100.0}\% & 85.0\% & 82.0\% & 89.34\% \\
    AWQ (Object Calib) & {91.5}\% & 98.5\% & {92.0}\% & 78.0\% & 90.00\% \\
    AWQ (Goal Calib) & 92.5\% & \textbf{100.0}\% & \textbf{93.5}\% & 84.5\% & 92.63\% \\
    AWQ (10 Calib) & \underline{94.0}\% & \underline{99.5}\% & \underline{92.5}\% & 76.5\% & 90.63\% \\
    \rowcolor{lgreen}
    \textbf{TTQ} ($r=0$) &
    {93.0}\% & \underline{99.5}\% & \textbf{93.5}\% & \underline{87.0}\% & \underline{93.25}\% \\
    \rowcolor{lgreen}
    \textbf{TTQ} ($r=16$) &
    \textbf{94.5}\% & \textbf{100.0}\% & \textbf{93.5}\% & \textbf{87.5}\% & \textbf{93.88}\% \\
    \bottomrule
    \end{tabular}
\end{table}

\section{LLM Models}
\label{sec:model}

\paragraph{OPT Family}
The Open Pre-trained Transformers (OPT)~\citep{zhang2022opt} is a suite of decoder-only pre-trained transformers ranging from 125M to 175B parameters. 
It was claimed that OPT-175B is comparable to GPT-3, while requiring only 1/7th the carbon footprint to develop.
Table~\ref{tab:opt} shows model parameters for the OPT open LLM family.

\paragraph{Qwen3 Family}
We use Qwen3~\citep{yang2025qwen3} dense models, which are decoder-only transformers spanning 270M to 30B parameters, built on a consistent architecture with RMSNorm, SwiGLU feed-forward layers, and rotary positional embeddings. 
All variants employ grouped-query attention (GQA) with a fixed small number of key–value heads while scaling the number of query heads with model width, reducing KV-cache cost. 
Importantly, the hidden size is decoupled from the attention projection width, providing additional flexibility.
Parameters are listed in Table~\ref{tab:qwen3-dense}.

\paragraph{Gemma3 Family}
Gemma3 models~\citep{team2025gemma} are decoder-only transformer architectures released across a wide range of scales, from 270M to 27B parameters, and include both text-only and multimodal variants. 
Similar to Qwen3, all Gemma3 models adopt RMSNorm, SwiGLU feed-forward networks, and rotary positional embeddings, as well as grouped-query attention (GQA). 
The per-head dimension is fixed at 256 across model sizes. 
Parameters are listed in Table~\ref{tab:gemma3-it}.

\begin{table}[t]
\centering
\caption{OPT model parameters~\cite{zhang2022opt}}
\label{tab:opt}
\small
\begin{tabular}{rrrrrrl}
  \toprule
  Model & layers &  heads  & hidden size & head dim & MLP dim & Huggingface ID \\
  \midrule
  125M   & 12 & 12 & 768 & 64 & 3072 & \href{https://huggingface.co/facebook/opt-125m}
  {facebook/opt-125m} \\
  350M   & 24 & 16 & 1024 & 64 & 4096 & \href{https://huggingface.co/facebook/opt-350m}
  {facebook/opt-350m} \\
  1.3B   & 24 & 32 & 2048 & 64 &8192 & \href{https://huggingface.co/facebook/opt-1.3b}
  {facebook/opt-1.3b}\\
  2.7B   & 32 & 32 & 2560 & 80 &10240 & \href{https://huggingface.co/facebook/opt-2.7b}
  {facebook/opt-2.7b}\\
  6.7B   & 32 & 32 & 4096 & 128 &16384 & \href{https://huggingface.co/facebook/opt-6.7b}
  {facebook/opt-6.7b}\\
  13B   & 40 & 40 & 5120 & 128 &20480 & \href{https://huggingface.co/facebook/opt-13b}
  {facebook/opt-13b} \\
  30B   & 48 & 56 & 7168 & 128&28672 & \href{https://huggingface.co/facebook/opt-30b}
  {facebook/opt-30b} \\
  66B   & 64 & 72 & 9216 & 128&36864 & \href{https://huggingface.co/facebook/opt-66b}
  {facebook/opt-66b} \\
  \bottomrule
\end{tabular}
\end{table}

\begin{table}[t]
\centering
\caption{Architecture parameters of Qwen3 dense models~\cite{yang2025qwen3}}
\label{tab:qwen3-dense}
\small
\begin{tabular}{rrrrrrrl}
\toprule
Model
& layers
& heads
& KV heads
& hidden size
& head dim
& MLP dim 
& Huggingface ID \\
\midrule
0.6B
& 28 & 16 & 8
& 1024 & 128 
& 3072 &
\href{https://huggingface.co/Qwen/Qwen3-0.6B}
  {Qwen/Qwen3-0.6B}\\

1.7B
& 28 & 16 & 8
& 2048 & 128 
& 6144 &
\href{https://huggingface.co/Qwen/Qwen3-1.7B}
  {Qwen/Qwen3-1.7B}\\

4B
& 36 & 32 & 8
& 2560 & 128 
& 9728 &
\href{https://huggingface.co/Qwen/Qwen3-4B}
  {Qwen/Qwen3-4B}\\

8B
& 36 & 32 & 8
& 4096 & 128 
& 12288 &
\href{https://huggingface.co/Qwen/Qwen3-8B}
  {Qwen/Qwen3-8B}\\

14B
& 40 & 40 & 8
& 5120 & 128 
& 17408 &
\href{https://huggingface.co/Qwen/Qwen3-14B}
  {Qwen/Qwen3-14B}\\

32B
& 64 & 64 & 8
& 5120 & 128 
& 25600 &
\href{https://huggingface.co/Qwen/Qwen3-32B}
  {Qwen/Qwen3-32B}\\
\bottomrule
\end{tabular}
\end{table}

\begin{table}[t]
\centering
\caption{Gemma3 instruction-tuned text transformer parameters~\cite{team2025gemma}}
\label{tab:gemma3-it}
\small
\begin{tabular}{rrrrrrrl}
\toprule
Model
& layers
& heads
& KV heads
& hidden size
& head dim
& MLP dim
& Huggingface ID \\
\midrule
270M
& 18 & 4 & 1
& 640 & 256 & 2048
& \href{https://huggingface.co/google/gemma-3-270m-it}
  {google/gemma-3-270m-it} \\

1B
& 26 & 4 & 1
& 1152 & 256 & 6912
& \href{https://huggingface.co/google/gemma-3-270m-it}
  {google/gemma-3-1b-it} \\

4B
& 34 & 8 & 4
& 2560 & 256 & 10240
& \href{https://huggingface.co/google/gemma-3-270m-it}
  {google/gemma-3-4b-it} \\

12B
& 48 & 16 & 8
& 3840 & 256 & 15360
& \href{https://huggingface.co/google/gemma-3-270m-it}
  {google/gemma-3-12b-it} \\

27B
& 62 & 32 & 16
& 5376 & 256 & 21504
& \href{https://huggingface.co/google/gemma-3-270m-it}
  {google/gemma-3-27b-it} \\
\bottomrule
\end{tabular}
\end{table}

\paragraph{LLaVA Family}
LLaVA~\cite{liu2023LLaVA} integrates a vision encoder with LLMs to enable processing both language and visual context modalities through instruction-tuning. 
The vision encoder typically uses contrastive language-image pretraining (CLIP) vision transformer (ViT), while the LLMs are based on LLaMA or Vicuna.
It was claimed that LLaVA-1.6-34B outperforms Gemini Pro on some benchmarks including MMMU and MathVista.

\paragraph{Qwen3-VL Family}
Qwen3-VL~\cite{bai2025qwen3} is a family of multimodal LLMs that extend the Qwen3 transformer with vision–language capabilities. 
The models integrate a SigLIP-based vision encoder with the Qwen3 language backbone through a projection module, enabling joint reasoning over images, videos, and text for tasks such as visual question answering, captioning, and document understanding.
Parameters are listed in Table~\ref{tab:qwen3vl}.

\begin{table}[t]
\centering
\caption{Qwen3-VL transformer parameters~\cite{bai2025qwen3}}
\label{tab:qwen3vl}
\small
\setlength{\tabcolsep}{4pt}
\begin{tabular}{rrrrrrrl}
\toprule
Model
& layers
& heads
& KV heads
& hidden size
& head dim
& MLP dim
& Hugginface ID
\\
\midrule
2B	& 24 & 16 & 8 &
2048 & 128 & 11008 &
\href{https://huggingface.co/Qwen/Qwen3-VL-2B-Instruct}
  {Qwen/Qwen3-VL-2B-Instruct}
\\
4B & 32 & 32 & 8 &
4096 & 128 & 22016 &
\href{https://huggingface.co/Qwen/Qwen3-VL-4B-Instruct}
  {Qwen/Qwen3-VL-4B-Instruct}
\\
8B & 36 & 32 & 8 &
4096 & 128 & 22016 &
\href{https://huggingface.co/Qwen/Qwen3-VL-8B-Instruct}
  {Qwen/Qwen3-VL-8B-Instruct}
\\
32B & 64 & 40 & 8 &
5120 & 128 & 27392 &
\href{https://huggingface.co/Qwen/Qwen3-VL-32B-Instruct}
  {Qwen/Qwen3-VL-32B-Instruct}
\\
\bottomrule
\end{tabular}
\end{table}

\paragraph{$\boldsymbol{\pi_{0.5}}$ Family}

The $\pi_{0.5}$ model~\cite{intelligence2025pi05} is a PaliGemma-based state-of-the-art VLA transformer having 2.3B parameters, that maps visual observations and language instructions directly to continuous robot actions through flow-matching diffusion policy. 
Trained via distillation from a larger foundation model on diverse robot interaction data, it achieves strong zero-shot generalization while remaining lightweight and deployable for real-world manipulation tasks.
Parameters are listed in Table~\ref{tab:pi05}.

\begin{table}[t]
\centering
\caption{$\pi_{0.5}$ VLA transformer parameters~\cite{intelligence2025pi05}:
Huggingface ID \href{https://huggingface.co/lerobot/pi05_libero_finetuned}
  {lerobot/pi05\_libero\_finetuned}}
\label{tab:pi05}
\small
\begin{tabular}{lrrrrrr}
\toprule
Module
& layers
& heads
& KV heads
& hidden size
& head dim
& MLP dim
\\
\midrule
LLM: Gemma-2B
& 18 & 8 & 1
& 2048 & 256 & 16384
 \\

Vision Encoder: SigLIP ViT-L
& 24 & 16 & 16
& 1024 & 64 & 4096
\\
Diffusion Policy: Gemma-300M
&
18 & 8 & 1 &
1024 & 128 & 4096
\\
\bottomrule
\end{tabular}
\end{table}

\section{Datasets}
\label{sec:dataset}

\paragraph{Wikitext-2 (WT2)}
The WikiText language modeling dataset~\citep{merity2016pointer} is a collection of over 100 million tokens extracted from the set of verified good and featured articles on Wikipedia. 
The dataset is available under the CC BY-SA-4.0 license.
The wikitext-2-raw-v1 contains 36{,}718, 3{,}760, and 4{,}358 samples for train, validation, and test splits, respectively.
We use \url{https://huggingface.co/datasets/mindchain/wikitext2}.

\paragraph{Penn Treebank (PTB)}
The English Penn Treebank (PTB) corpus~\citep{marcus1994penn}
is one of the most known and used corpus for the evaluation of models for sequence labeling. 
The dataset features a million words of 1989 Wall Street Journal material. 
We use \url{https://huggingface.co/datasets/ptb-text-only/ptb_text_only}.

\paragraph{C4}
C4~\citep{raffel2020exploring} is based on a colossal, cleaned version of Common Crawl's web crawl corpus.
This is release under the OCD-By license.
We consider a subset ``en'', containing 364{,}868{,}892 and 364{,}608 samples for train and validation splits, respectively, while we use the first shard for each split in \url{https://huggingface.co/datasets/allenai/c4}.

\paragraph{ScienceQA}
ScienceQA~\cite{lu2022learn} is collected from elementary and high school science curricula (i.e., grades 1 through 12), and contains 21{,}208 multimodal multiple-choice science questions. 
Out of the questions in ScienceQA, 10{,}332 (48.7\%) have an image context, 10{,}220 (48.2\%) have a text context, and 6{,}532 (30.8\%) have both. 
Most questions are annotated with grounded lectures (83.9\%) and detailed explanations (90.5\%). 
The lecture and explanation provide general external knowledge and specific reasons, respectively, for arriving at the correct answer. 
ScienceQA has rich domain diversity from three subjects: natural science, language science, and social science. 
ScienceQA features 26 topics, 127 categories, and 379 skills that cover a wide range of domains.
We use \url{https://huggingface.co/datasets/derek-thomas/ScienceQA},  released under the CC BY-NC-SA 4.0 license.

\paragraph{TextVQA}

TextVQA~\cite{singh2019towards} requires VLM models to read and reason about text in images to answer questions about them. 
Specifically, models need to incorporate the new modality of text present in the images and reason over it to answer TextVQA questions. 
TextVQA dataset contains 45{,}336 questions over 28{,}408 images from the OpenImages dataset. 
We use \url{https://huggingface.co/datasets/llms-eval/textvqa}, licensed under CC-BY-4.0.

\paragraph{COCO-Caption}
The COCO-Caption dataset is a standard benchmark for image captioning built on the Microsoft COCO dataset~\cite{lin2014microsoft}. 
It contains over 120k images, each annotated with five human-written captions describing the visual scene. 
The images cover diverse everyday environments with multiple objects and interactions for evaluating VLMs.
We use \url{https://huggingface.co/datasets/llms-lab/COCO-Caption2017}, licensed under CC-BY-4.0.

\paragraph{OK-VQA}
The OK-VQA dataset~\cite{marino2019ok} is a benchmark for knowledge-based visual question answering, where answering questions requires external world knowledge beyond the visual content of the image. It contains over 14k questions paired with images from the MS COCO dataset, along with multiple human-provided answers for evaluation. The questions cover diverse topics such as science, history, and common knowledge, making the task more challenging than standard VQA benchmarks. 
We use \url{https://huggingface.co/datasets/llms-lab/COCO-Caption2017}, inheriting the COCO image license of CC-BY-4.0, while the annotations are released for research use.

\paragraph{ChartQA}
The ChartQA dataset~\cite{masry2022chartqa} is a benchmark for chart-based visual question answering, designed to evaluate models’ ability to interpret charts and perform numerical and logical reasoning. It contains 33k question–answer pairs over 21k charts collected from multiple real-world sources. 
The dataset includes both human-written and automatically generated questions, requiring models to extract information or compute values from chart data. We use \url{https://huggingface.co/datasets/llms-lab/ChartQA}, released under the GPL-3.0 license.

\paragraph{LIBERO}
The LIBERO dataset~\cite{liu2023libero} is a benchmark for robotic vision-language-action (VLA) learning that evaluates long-horizon, compositional manipulation in simulated household environments. It provides diverse task suites with structured train/test splits designed to measure cross-task generalization, skill composition, and transfer to novel object and scene configurations. Each task includes multimodal data---video observations, language instructions, and low-level control trajectories---enabling end-to-end learning from vision and language to robot actions.
We use \url{https://huggingface.co/datasets/lerobot/libero},
 licensed under Apache-2.0.

\end{document}